\documentclass[sigconf,screen]{acmart}

\usepackage{multirow}
\usepackage{wrapfig}
\usepackage{xspace}
\usepackage[capitalise,noabbrev,nameinlink]{cleveref}
\crefformat{equation}{#2Equation~#1#3}
\usepackage{pgfplots,pgfplotstable}
\usepackage{subcaption}
\usepgfplotslibrary{colorbrewer,groupplots}
\usetikzlibrary{positioning}
\pgfplotsset{compat=1.18}
\newcommand{\eg}{e.g.\@\xspace}
\newcommand{\ie}{i.e.\@\xspace}
\newcommand{\bb}[1]{\textcolor{blue!80!black}{\underline{#1}}}

\newcommand{\ceil}[1]{\left\lceil#1\right\rceil}
\newcommand{\floor}[1]{\left\lfloor#1\right\rfloor}
\newcommand{\nits}{cd/m\textsuperscript{2}\xspace}
\DeclareMathOperator{\PQ}{PQ}

\copyrightyear{2023}
\acmYear{2023}
\setcopyright{rightsretained}
\acmConference[SA Conference Papers '23]{SIGGRAPH Asia 2023 Conference Papers}{December 12--15, 2023}{Sydney, NSW, Australia}
\acmBooktitle{SIGGRAPH Asia 2023 Conference Papers (SA Conference Papers '23), December 12--15, 2023, Sydney, NSW, Australia}
\acmDOI{10.1145/3610548.3618139}
\acmISBN{979-8-4007-0315-7/23/12}

\usepackage{xstring}
\newcommand{\showDOI}[1]{%
  \let\oldleft\UrlLeft%
  \let\oldright\UrlRight%
  \def\UrlLeft##1\UrlRight{\StrGobbleLeft{##1}{16}}%
  \textsc{doi:} #1%
  \let\UrlLeft\oldleft%
  \let\UrlRight\oldright%
}

\graphicspath{{figures/}}

\newcommand{\mainorsup}[2]{{#1}}

\begin{document}

\title{VR-NeRF: High-Fidelity Virtualized Walkable Spaces}

\author{Linning Xu}
\orcid{0000-0003-1026-2410}
\affiliation{\institution{The Chinese University of Hong Kong}\country{Hong Kong}}
\affiliation{\institution{Meta}\country{USA}}

\author{Vasu Agrawal}
\orcid{0000-0003-3077-1212}
\affiliation{\institution{Meta}\country{USA}}

\author{William Laney}
\orcid{0009-0006-8939-0106}
\affiliation{\institution{Meta}\country{USA}}

\author{Tony Garcia}
\orcid{0009-0001-0670-3343}
\affiliation{\institution{Meta}\country{USA}}

\author{Aayush Bansal}
\orcid{0000-0002-0966-3930}
\affiliation{\institution{Meta}\country{USA}}

\author{Changil Kim}
\orcid{0000-0002-6541-8825}
\affiliation{\institution{Meta}\country{USA}}

\author{Samuel Rota Bulò}
\orcid{0000-0002-2372-1367}
\affiliation{\institution{Meta}\country{Switzerland}}

\author{Lorenzo Porzi}
\orcid{0000-0001-9331-2908}
\affiliation{\institution{Meta}\country{Switzerland}}

\author{Peter Kontschieder}
\orcid{0000-0002-9809-664X}
\affiliation{\institution{Meta}\country{Switzerland}}

\author{Aljaž Božič}
\orcid{0009-0002-2985-6921}
\affiliation{\institution{Meta}\country{Switzerland}}

\author{Dahua Lin}
\orcid{0000-0002-8865-7896}
\affiliation{\institution{The Chinese University of Hong Kong}\country{Hong Kong}}

\author{Michael Zollhöfer}
\orcid{0000-0003-1219-0625}
\affiliation{\institution{Meta}\country{USA}}

\author{Christian Richardt}
\orcid{0000-0001-6716-9845}
\affiliation{\institution{Meta}\country{USA}}

\renewcommand{\shortauthors}{Xu et al.}

\begin{abstract}
We present an end-to-end system for the high-fidelity capture, model reconstruction, and real-time rendering of walkable spaces in virtual reality using neural radiance fields.
To this end, we designed and built a custom multi-camera rig to densely capture walkable spaces in high fidelity and with multi-view high dynamic range images in unprecedented quality and density.
We extend instant neural graphics primitives with a novel perceptual color space for learning accurate HDR appearance, and an efficient mip-mapping mechanism for level-of-detail rendering with anti-aliasing, while carefully optimizing the trade-off between quality and speed.
Our multi-GPU renderer enables high-fidelity volume rendering of our neural radiance field model at the full VR resolution of dual 2K$\times$2K at 36\,Hz on our custom demo machine.
We demonstrate the quality of our results on our challenging high-fidelity datasets, and compare our method and datasets to existing baselines.
We release our dataset on our project website: \url{https://vr-nerf.github.io}.
\end{abstract}


\begin{teaserfigure}
    \centering
    \includegraphics[width=\textwidth]{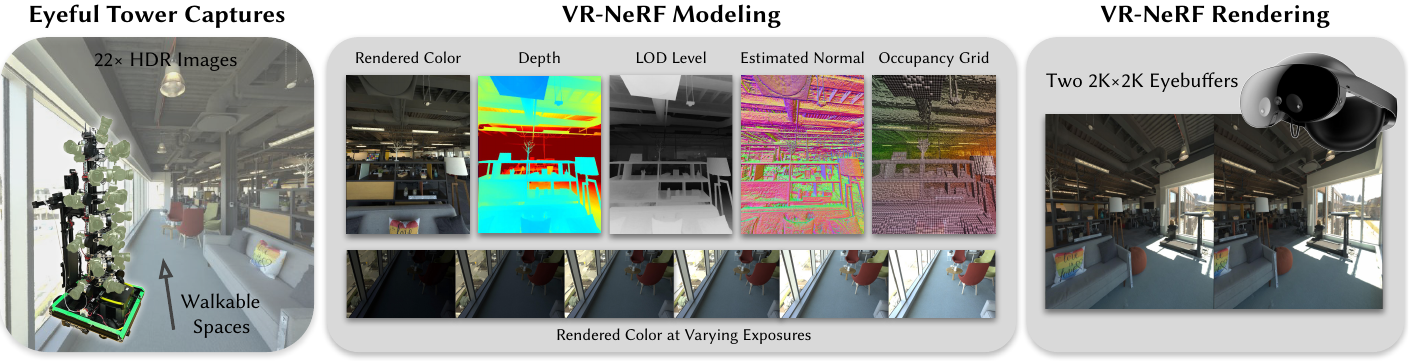}
    \caption{\label{fig:teaser}%
        VR-NeRF brings high-fidelity walkable spaces to real-time virtual reality.
        Our ``Eyeful Tower'' camera rig captures spaces with high image resolution and dynamic range that approach the limits of the human visual system.
        We train high-fidelity neural radiance fields that exploit the high-dynamic range nature of our captured scenes and provide level-of-detail mip-mapping for efficient anti-aliasing.
        Our rendering backend leverages our accurate occupancy grid and a dynamic multi-GPU work distribution scheme to achieve real-time frame rates on dual 2K$\times$2K eyebuffers for an immersive VR experience.
        \\
    }
\end{teaserfigure}

\maketitle

\section{Introduction}
\label{sec:introduction}

The advent of consumer virtual reality (VR) headsets has led to a proliferation of highly immersive visual media, including breathtaking VR photography and video.
However, existing approaches support either high-fidelity view synthesis with a small \emph{headbox} of less than 1\,m diameter \cite{BroxtFOEHDDBWD2020,OverbEEPD2018}, or scene-scale free-viewpoint view synthesis of lower quality or framerate \cite{ParraTFHMSSC2019,JangMKKRK2022,WuXZBHTX2022}.
In this work, we present a comprehensive system designed to overcome these limitations all the way from capture to rendering for high-fidelity free-viewpoint exploration of walkable, real-world static spaces in VR.
Our contributions address the following challenges:
\begin{enumerate}
	\item dense, high-fidelity capture of large-scale walkable spaces,
	\item high-fidelity neural radiance field reconstruction, and
	\item real-time rendering of our neural radiance fields in VR.
\end{enumerate}

\noindent
High-fidelity view synthesis depends on high-quality, densely captured multi-view images.
While NeRF objects use 100s of views \cite{MildeSTBRN2020} and light field captures around 1,000 views per location \cite{BroxtFOEHDDBWD2020}, walkable scenes will need a minimum of several thousand input views to provide enough spatial coverage.
Existing captures of walkable spaces tend to be handheld and usually comprise 100s of photos \cite[e.g.][]{PhiliMGD2021} or 1,000s of video frames \cite[e.g.][]{KnapiPZK2017}.
In both cases, the space of camera poses is undersampled: photo sequences lack sufficient density, and videos move along a 1D subspace that fails to sample the 6D pose space sufficiently uniformly.
High-fidelity view synthesis also needs to reproduce the high dynamic range of the real world, which existing methods do not.
To this end, we designed a custom camera rig that enables capturing walkable spaces in unprecedented quality and density: our datasets contain thousands of 50\,megapixel high dynamic range (HDR) images.
Several of our datasets exceed 100\,gigapixels -- two orders of magnitude more than existing datasets \cite{FlynnBDDFOST2019,PhiliMGD2021,XuWZHYBX2021}.

Neural radiance fields (NeRFs) have led to an explosion in high-quality novel-view synthesis techniques \cite{MildeSTBRN2020,TewarTMSTWLSMLSTNBWZG2022}.
However, existing methods do not support the size, scale, and dynamic range of our high-fidelity datasets, even when downsampled to 2K resolution.
We propose VR-NeRF, which is uniquely adapted to our high-quality datasets and supports real-time VR rendering in full NeRF quality.
Specifically, we introduce a new perceptually based color space for representing high-dynamic range radiance values of up to 10,000\,cd/m\textsuperscript{2}, allowing our model to learn up to 22 stops\footnote{One stop is a doubling or halving of the amount of light reaching the imaging sensor.} of dynamic range (or 4,194,304:1).
A second crucial component is a real-time-capable mip-mapping technique that suppresses aliasing when observing objects at different distances using level-of-detail rendering.
We also developed a principled pruning stage to obtain an accurate occupancy grid for speeding up rendering with a focus on improved geometry estimation.

The third and final stage of our end-to-end system is a custom multi-GPU renderer that brings high-fidelity NeRF rendering into virtual reality.
On our custom-built demo machine, we can render our models at the full resolution of the Quest Pro VR headset, \ie, two 2K$\times$2K eye buffers (\textasciitilde8\,megapixel), at a consistent frame rate of 36\,Hz, which results in a compelling VR experience that enables free exploration of walkable spaces in high fidelity.

\section{Related Work}
\label{sec:related-work}

\citet{KanadNR1995} coined the term “Virtualized Reality” to see a previously recorded event from any perspective.
Our goal is to virtually walk through previously captured scenes at high fidelity in virtual reality.
We, therefore, call our work \emph{Virtualized Walkable Spaces}.
There are three crucial components to enable high-fidelity virtualized walkable spaces:
(1) a mobile high-resolution multi-view camera system to densely capture large-scale scenes;
(2) an efficient neural representation to compactly and accurately encode a large-scale scene with high dynamic range and level of detail; and
(3) optimized real-time rendering at VR resolution and frame rate.

\paragraph{High-Resolution Multi-View Capture System}
Capture systems can vary from a single moving camera \cite{GortlGSC1996,LevoyH1996,DavisLD2012,KimZPSG2013,KnapiPZK2017,HedmaRDB2016,BerteYLR2020} to multi-camera rigs \cite{WilbuJVTABAHL2005,BroxtFOEHDDBWD2020,FlynnBDDFOST2019,ParraTFHMSSC2019} and synchronized camera arrays in big studios \cite{JooSLLTGBGNMKNS2019,OrtsERFCKDKDKDTLCCMVPWKKLKI2016}.
Existing multi-view captures are either limited to a small headbox \cite[e.g.][]{OverbEEPD2018,ParraTFHMSSC2019} or are sparsely captured \cite[e.g.][]{YoonKGPK2020,KnapiPZK2017}, which restricts freedom of motion.
We built a multi-camera rig that densely and efficiently captures a wide variety of walkable spaces to create large-scale multi-view datasets with high-resolution details (50\,megapixels) and high dynamic range.

\paragraph{Large-scale Novel View Synthesis}
Our focus is on real-time VR rendering of high-fidelity walkable spaces; recent surveys cover the full range of scene representations \cite{TewarTMSTWLSMLSTNBWZG2022,RichaTW2020}.
While mesh-based reconstructions \cite{WhelaGLSGSBVN2018,StrauWMCWGEMRVCYBYPYZLCBGMPSBSNGLN2019} are ideal for fast rendering, they tend to lack fine geometric detail.
Image-based rendering \cite[e.g.][]{HedmaRDB2016} achieves more visual detail but struggles with reflective surfaces.
Several follow-up methods use neural representations for explicit reflection support \cite{WuXZBHTX2022,PhiliMGD2021,XuWZHYBX2021} and achieve interactive frame rates.
NeRFs \cite{MildeSTBRN2020} have become the de-facto standard neural representation due to their versatility and ability to represent complex scenes with high fidelity.
They have been extended in multiple ways to represent large-scale scenes even at a city scale
\cite{TanciCYPMSBK2022,TurkiRS2022,XiangXPZRTDL2022,XuXPPZTDL2023,ZhangCC2023}.
High-resolution concerns have also been addressed \cite{WangLSSB2022, JiangHMXBWX2023}.
However, these methods do not support level of detail and high dynamic range, which are required for high-fidelity VR.
LocalRF \cite{MeuleLGHKKK2023} and F\textsuperscript{2}-NeRF \cite{WangLCLLKTW2023} tackle large unbounded scenes, yet only support limited view extrapolation and thus cannot provide fully immersive free-view exploration.
Methods built on implicit surfaces, like signed distance functions, tend to focus on high-quality 3D surface reconstruction rather than view synthesis \cite{YuPNSG2022,ZhuHYLLXWTHBW2023,LiMETULL2023,RosuB2023}.
We build our model on Instant-NGP (iNGP) \cite{MuelleESK2022}, as it supports real-time rendering without model baking, and provides easily extensible model capacity via its hash grid.
However, it lacks support for high-fidelity rendering of large-scale walkable spaces, such as level of detail and perceptually based HDR support.
\looseness-1

\paragraph{High Dynamic Range (HDR)}
The human visual system supports a significantly higher dynamic range than current camera or display technology \cite{ReinhWPD2006}.
When recreating highly realistic walkable spaces, it is therefore important to accurately capture and render the scene in HDR.
RawNeRF learns linear radiance from raw sensor measurements using a weighted L2 loss that approximates a tonemapped loss \cite{MildeHMSB2022}.
Several methods learn to reconstruct linear radiance from low dynamic range images using differentiable tonemapping models \cite{RueckeFS2022,HuangZFLWW2022,JunSeYYO2022}.
We train our HDR model directly using HDR input images in a novel perceptually uniform color space that does not require custom losses or tonemapping modules.

\paragraph{Level of Detail (LOD)}
\citet{TakikLYKLNJMF2021} and \citeauthor{BarroMTHMS2021}'s Mip-NeRF \citeyearpar{BarroMTHMS2021} introduced the notion of level of detail into neural signed distance and radiance fields, respectively, to reduce geometric and visual complexity, e.g. to minimize aliasing when viewing objects from a distance.
As Mip-NeRF's integrated positional encoding is incompatible with efficient grid-based NeRF approaches like iNGP \cite{MuelleESK2022}, Zip-NeRF \cite{BarroMVSH2023} uses supersampling as an approximation, but multi-second inference times still prevent real-time rendering.
\citet{AroudLISGN2022} store the scene redundantly at multiple LOD levels in a sparse voxel octree.
We introduce an efficient LOD approach designed for iNGP that enables high-fidelity real-time VR rendering with anti-aliasing.

\section{The ``Eyeful Tower'' Capture Rig}
\label{sec:eyeful_tower}

Capturing scenes with a hand-held camera quickly reaches limits: taking hundreds of photos is tedious, achieving consistent coverage of viewpoints is difficult, and hand-held exposure bracketing is tricky due to camera shake.
To capture real-world environments with the highest visual fidelity in terms of spatial resolution and dynamic range, we designed, built, and refined a custom multi-camera capture rig affectionately referred to as the \emph{Eyeful Tower}.
The design of our capture rig was guided by the following considerations:

\begin{enumerate}
\setlength{\itemsep}{0.15em}
	
	\item \emph{Coverage:}
	Place cameras for approximately uniform light field capture, and parallelize data capture across cameras.
	
	\item \emph{Fidelity:}
	Match human visual perception in terms of acuity and high dynamic range.
	
	\item \emph{Mobility:}
	Allow single-person operation, and be usable without external power or network connection.
	
	\item \emph{Rigidity:} Support multi-exposure bracketing for high dynamic range (HDR) reconstruction without camera motion.
	
	\item \emph{Storage:} Record photos on-camera, so no server is needed. Offload all photos via a single network cable.
	
\end{enumerate}

\subsection{Capture Rig Design}

\setlength{\columnsep}{5pt}
\begin{wrapfigure}[14]{r}{95pt}
\vspace{-2.75\baselineskip}
\includegraphics[width=\linewidth]{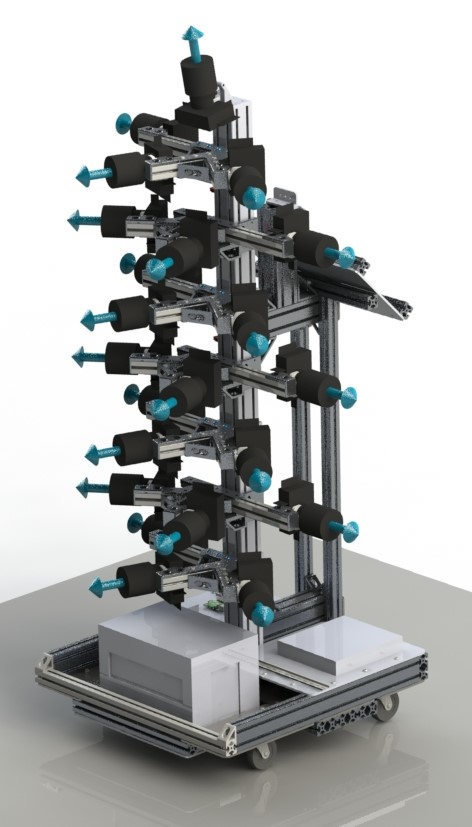}
\end{wrapfigure}
We built our capture rig using extruded aluminium around an 80$\times$80\,cm base with a 1.8\,m vertical pole for 22 cameras that are distributed on 7 levels with 3 cameras each, plus one upward-facing camera at the top (see right).
A 1.5\,kWh Li-ion battery powers cameras, a 24-port network switch, and a Raspberry Pi controlling the cameras.
We chose Sony $\alpha$1 mirrorless cameras for their high-quality 50-megapixel raw images with 14 stops of dynamic range.
Please see our supplement for details on the rig design and camera/lens choices.

\subsection{Capture Process}

\subsubsection{Desirable Capture Density}\label{sec:desirable-capture-density}

Reproducing the appearance of a static scene from any viewpoint in theory requires observations for the entire 5D plenoptic function \cite{AdelsB1991}.
The widely used NeRF synthetic dataset \cite{MildeSTBRN2020} has viewpoints densely distributed on a hemisphere, which allows the renderings to generalize continuously across the whole hemisphere of viewing directions.
For scene-scale rendering, we are lacking such a densely captured dataset, which results in the \emph{limited capability to extrapolate novel viewpoints}.
However, this is critically important for virtual reality, where we want to deliver walkable spaces with 6-degrees-of-freedom allowable head movement.

\subsubsection{Capture Procedure}\label{sec:capture-procedure}

We capture scenes by `tiling' the available floor area with rig positions that are spaced roughly 30\,cm apart.
For complete captures, we capture forward- and backward-facing views, while trying to stay at least 30\,cm away from walls or objects.
Near walls, it is often sufficient to only capture the direction facing away from the wall, as defocused close-ups of a wall usually add little value.
Before each capture, we also place scale bars (for automatic scale estimation) and a Macbeth ColorChecker (for color verification and white balance) into the scene.
During the capture, we try to stay out of view of any camera, avoid moving any objects, such as chairs or carpets, and aim to minimize lighting changes and shadow casting.
\looseness-1

\subsection{Data Preprocessing}

\subsubsection{HDR Image Merging}

We use LibRaw 0.21 to debayer the raw images captured by our cameras to 16-bit linear TIFF images.
We then merge 9 different exposures into one high-dynamic range image using a robustified version of \citeauthor{HanjiZM2020}'s Poisson photon noise estimator \citeyearpar{HanjiZM2020}, which provides an unbiased estimate of scene radiance.
We observed that the Sony $\alpha$1 raw image values do not saturate as quickly as expected, which produces outliers that can reduce the estimated radiance sufficiently to cause visible color changes.
Therefore, we keep track of the minimum radiance estimate per pixel and color channel, so that we can ignore it when merging the input exposures.
In addition, we set fully saturated pixels to the lowest radiance that saturates in all images.

\subsubsection{Camera Calibration}

We estimate camera poses and intrinsics using Agisoft Metashape Pro 2.0 \cite{Agiso2023}, a professional photogrammetry software that supports rig calibration, in which the relative pose between cameras is constant across all positions of the capture rig within a scene.
Metashape effectively handles our large-scale datasets with up to 6,300 photos at 50\,megapixel resolution \cite{OverRKBBNSWW2021}.
It also automatically detects the markers on our calibrated scale bars, such that camera poses are in metric space for 1:1 scale rendering in VR.

\subsubsection{Captured Datasets}

We captured multiple datasets using our Eyeful Tower capture rig, which are summarized in \cref{tab:dataset_stats}.
Our captures took between 5\,minutes and 6\,hours, depending on the scale and complexity of the scene, with an average speed of around one minute per m\textsuperscript{2}.
The resulting datasets comprise 29–303 billion pixels, or rays, covering spaces of 6–120\,m\textsuperscript{2}.

\begin{table}[hb]
    \caption{\label{tab:dataset_stats}%
        Statistics of scenes captured using our Eyeful Tower rig:
        We show the number of cameras, rig positions, and images, as well as the capture time, surface area, and the number of rays at full resolution (5,784$\times$8,660) and 1368$\times$2048 (`2K'), our typical training and rendering image resolution.
    }
    \centering\sffamily
    \def\sqm{m\textsuperscript{2}}%
    \renewcommand*{\arraystretch}{1.15}%
    \resizebox{0.96\linewidth}{!}{%
        \begin{tabular}{lrrrrrrr}
        \toprule
        Scene                  & \hspace{-2.5em}Cameras &  \# Pos. &  \# Img.  &      Time    &     Area     &    Rays   &   Rays @ 2K     \\
        \midrule
        \textsc{apartment}     &    22   &    180   &   3,960   &    60\,min   &   55\,\sqm   &  190.6\,B  & 10.7\,B   \\
        \textsc{kitchen}       &    19   &    318   &   6,024   &    43\,min   &   54\,\sqm   &  302.7\,B  & 16.9\,B   \\
        \textsc{office1a}      &     9   &     85   &     765   &    23\,min   &   20\,\sqm   &   29.1\,B  &  1.6\,B   \\
        \textsc{office1b}      &    22   &     71   &   1,562   &    16\,min   &   20\,\sqm   &   78.2\,B  &  4.4\,B   \\
        \textsc{office2}       &     9   &    233   &   2,097   &    39\,min   &   35\,\sqm   &   79.8\,B  &  4.5\,B   \\
        \textsc{office\_view1} &    22   &    126   &   2,772   &    31\,min   &   18\,\sqm   &  138.9\,B  &  7.8\,B   \\
        \textsc{office\_view2} &    22   &     67   &   1,474   &    10\,min   &   33\,\sqm   &   73.8\,B  &  4.1\,B   \\
        \textsc{riverview}     &    22   &     48   &   1,008   &     5\,min   &    6\,\sqm   &   52.9\,B  &  3.0\,B   \\
        \textsc{seating\_area} &     9   &    168   &   1,512   &    22\,min   &   16\,\sqm   &   55.9\,B  &  3.1\,B   \\
        \textsc{table}         &     9   &    134   &   1,206   &    14\,min   &   24\,\sqm   &   45.2\,B  &  2.5\,B   \\
        \textsc{workshop}      &     9   &    700   &   6,300   &   364\,min$^\dagger$\hspace{-4pt} & 120\,\sqm & 239.4\,B & 13.4\,B   \\
        \bottomrule
        \end{tabular}}
        \\[0.25em]\scriptsize $\dagger$ Includes 121\,minutes of capture time and 243\,minutes of data offload mid-capture.
\end{table}

\section{High-Fidelity Neural Radiance Fields}
\label{sec:method}

Volume rendering using neural radiance fields is a compelling choice for photorealistic scene representations due to the versatility of representing semi-transparent surfaces and finely detailed objects while being suitable for delivering scene-scale rendering.
As our focus is on maximizing rendering fidelity in the available compute budget, we use neural radiance fields as a foundation and leave alternative representations as future work.
Instead of constructing a large, complex model with extra capacity to account for various effects, our goal is to design a simple yet general model that facilitates real-time VR rendering for large-scale scenes.

We therefore build on the Instant NGP architecture \cite{MuelleESK2022} with its efficient and \emph{scalable} multi-level hash encoding for \emph{fast} rendering of large-scale static scenes.
We make several contributions to improve the visual fidelity of high-resolution room-scale rendering, including a perceptual color space that enables perceptual optimization of high dynamic range images using a simple $L_1$ loss.
We further introduce an efficient and effective level-of-detail scheme for anti-aliasing using multi-level hash grids.
To faithfully represent unbounded areas, such as views through windows or long corridors, we adopt a cubic space contraction
based on the $L_{\infty}$ norm \cite{WanRBLRNXLZRL2023}, which is a good fit for grid-based representations.
We discuss implementation details and additional components that contribute to the high quality of our view synthesis model in our supplement.
\looseness-1

\subsection{Perceptual Modeling of High Dynamic Range}
\label{sec:hdr}

\begin{figure}[h]
\includegraphics[width=\linewidth]{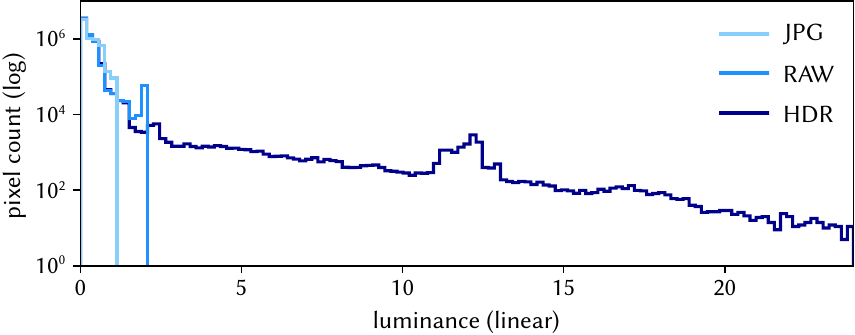}
\caption{\label{fig:hdr-histograms}%
    Comparison of the dynamic range of a JPEG photo (range 0\textasciitilde1) with the corresponding raw image (0\textasciitilde2)  and the full HDR image (0\textasciitilde145).}
\end{figure}

Our Sony $\alpha$1 cameras capture raw images with a dynamic range of 14 stops (\ie,~14\,bits of usable information).
The 9-step exposure bracketing adds a further 8 stops, for a total of 22 stops of dynamic range (see \cref{fig:hdr-histograms}).
In other words, the brightest input pixel value can be up to 4,194,304 times as bright as the darkest non-zero pixel value.
Applying common image losses like $L_1$ or $L_2$ directly in linear color spaces of this range leads to poor results as the losses are dominated by errors in bright areas.
For example, an error of 0.1 is significantly more noticeable at a base level of 0.1 (+100\%) compared to 10 (only +1\%), yet would be penalized the same.
The solution is to either use a more complex loss function, such as RawNeRF's relative MSE \cite{MildeHMSB2022}, or a carefully designed non-linear mapping to a perceptually uniform color space.

One such non-linear mapping is the Perceptual Quantizer (PQ) developed by Dolby \cite{MilleND2013} and standardized by \citet{SMPTE2084-2014}, which is the foundation of many consumer HDR image and video formats.
PQ was designed to optimally encode the large luminance range from 0 to 10,000\,\nits in 10–16\,bits while minimizing visible banding artifacts.
This was achieved by approximating the integral of just noticeable differences based on the contrast sensitivity function of the human visual system \cite{Kunke2022}.
The function
\begin{align}
\PQ(Y) &= \left(\frac{c_1 + c_2 \cdot Y^{m_1}}{1 + c_3 \cdot Y^{m_1}}\right)^{m_2} \quad \text{with constants}\\
m_1 &= \frac{1305}{8192} \text{, } m_2 = \frac{2523}{32} \text{, } c_1 = \frac{107}{128} \text{, } c_2 = \frac{2413}{128} \text{, } c_3 = \frac{2392}{128}
\end{align}
maps the input luminance $Y \!\in\! [0, 10{,}000]$\,\nits to the `PQ space' in the unit range.
For our experiments, we map linear color values of 1 to a luminance of 100\,\nits in order to allow a conversion to the PQ space.
Operating in the unit range is also a natural fit for the sigmoid activation function, which eases the learning of model output distributions.
Applying an $L_1$ or $L_2$ loss in the PQ space now penalizes errors according to human visual perception, and is able to produce colors in full high dynamic range.
See \cref{fig:exposure-swipe} for a sweep of different exposures at rendering time.

\subsection{Feature Grid Mip-Mapping for Level-of-Detail}
\label{sec:lod}

Level-of-detail (LOD) rendering is desirable for large-scale scenes, as objects observed at different distances reveal varying levels of geometric and texture detail.
Single-LOD methods like NeRF or iNGP can cause severe aliasing in highly textured objects seen at a distance, while details seen in only a few views might be washed out due to many overlapping distant views.
Multiple levels of detail can reduce aliasing as the LOD level can be dynamically adjusted based on the distance of objects from the viewer.
In computer graphics, texture LOD is usually implemented using mip-maps \cite{Willi1983}.
Mip-NeRF \cite{BarroMTHMS2021} introduced mip-mapping to NeRFs and Zip-NeRF \cite{BarroMVSH2023} recently extended these ideas to fast grid-based feature encodings, as used by iNGP.
Unfortunately, this approach is unsuitable for real-time rendering (1.1\,FPS on 8×V100).
Instead, we introduce a simple but effective mip-mapping scheme for grid-based feature encodings that enables learning of continuous LOD while actively supporting real-time rendering.

\subsubsection{Feature Grid Mip-Mapping}

Multi-resolution feature grids are a natural fit for LOD rendering as they already represent features across multiple scales.
By considering a ray as a cone as in Mip-NeRF, and by comparing its cross section with the size of grid features at each level, we can efficiently determine which feature grid levels are theoretically resolvable at the ray level, and can down-weight or even ignore finer levels that would introduce aliasing.

\begin{figure}
\includegraphics[width=\linewidth]{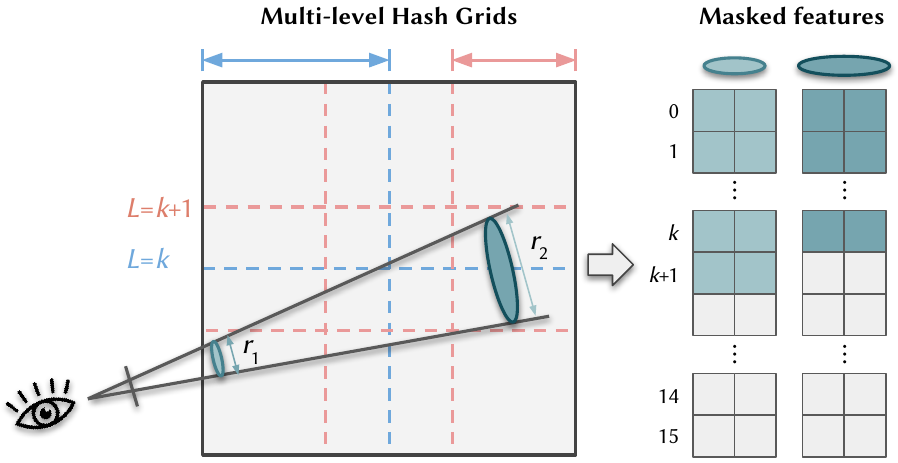}
\caption{\label{fig:lod}%
    \textbf{LOD Masking.}
    Smaller sample footprints like $r_1$ return more features from the multi-level hash grid than larger, more distant samples ($r_2$).
}
\end{figure}

For a specific ray, we start by calculating its base radius $r$ at unit distance along the ray.
At a sample location, the pixel footprint is then determined by multiplying the base radius with the metric distance $t$ along the ray as $\hat{r} = t \!\cdot\! r$.
For contracted spaces, \citet{BarroMVSH2022,BarroMVSH2023} and \citet{WangLCLLKTW2023} consider the Jacobian $\mathbf{J}_\mathcal{C}$ of the contraction function $\mathcal{C}(\cdot)$ at the sample location $\mathbf{x}$ to calculate the scale factor for variance or step size estimation.
Similarly, we could derive the contracted pixel radius via $\mathcal{C}(\hat{r}) = \hat{r} \cdot \sqrt[3]{\operatorname{det}\!\left(\mathbf{J}_\mathcal{C}(\mathbf{x})\right)}$.
In practice, we compute the contracted pixel radius directly from corresponding sample points on adjacent rays in the contracted space.
The optimal LOD level can then be calculated from the configuration of the multi-resolution feature grid as follows.
Suppose the base resolution is $s$ and the scale factor between levels is $f$.
For the $L$\textsuperscript{th} level (with $L=0$ being the base), each level has a grid voxel size of $(s f^L)^{-1}$.
Based on the Nyquist–Shannon sampling theorem, we dampen features whose size is less than twice the footprint $\hat{r}$ in the contracted coordinate space (see diagram in \cref{fig:lod}).
The optimal LOD level for a sample is therefore $L^* = - \log_f( 2s \hat{r})$.
For a piecewise linear LOD transition, we use these per-level weights:
\begin{align}
	w_L =
	\begin{cases}
		1 & L \leq \floor{L^*} \\
		L^* - \floor{L^*} & \floor{L^*} < L \leq \ceil{L^*} \\
		0 & \ceil{L^*} < L
	\end{cases}
\end{align}
For distant points, we only need to sample the features of the lowest few grid levels, which reduces rendering time, while gradually revealing high-frequency features for closer points.
Feature sampling of the finest hash grid layers is particularly expensive due to the highly incoherent memory access patterns.
Skipping these features results in substantially faster rendering (\cref{sec:rendering}).

\subsubsection{LOD Bias}

Similar to standard mip-mapping, we can optionally add an LOD bias $\Delta L$ to the LOD $L^*$ used for querying and weighting the grid features.
This continuously adjusts the sharpness of details to balance between blurred and aliased rendering.
In fact, the LOD bias can be viewed as a unifying framework that encompasses coarse-to-fine training strategies \cite{LinMTL2021,YangPW2023,ParkSBBGSB2021}.
Such progressive training approaches start with low-frequency models and gradually increase the number of feature scales to improve details.
This is equivalent to starting with a large negative LOD bias, such that only low-frequency features are used, and annealing it towards zero during training.
See \cref{fig:lod-swipe} for the visualized LOD bias sweep on the \textsc{apartment} scene.

\subsubsection{Distance-aware Features}
\label{sec:distance-aware-features}

By mip-mapping grid features, we are effectively making the features used for radiance computation distance-aware, as different features are used at different viewing distances.
This offers an additional degree of freedom to handle inconsistent data during the capture process, such as the distance-dependent shadows cast by the camera rig.
Rig shadows are most prominent when the rig approaches walls or corners.
With limited training views in these ambiguously captured locations, the model is likely to fake the shadows with incorrect geometry and/or appearance.
On the other hand, distance-aware features allow our model to learn distance-dependent appearance, which reduces visual artifacts.
We further noticed that the mip-mapped features encourage the model to better allocate model capacity for fine-grained details.

\subsection{Optimizing the Quality--Speed Trade-off}

Our goal of high-fidelity real-time NeRF rendering requires some challenging trade-offs between visual quality and rendering speed.
For example, while conditional latent codes and wider and deeper networks can improve rendering quality \cite{BarroMVSH2023,MuelleESK2022}, they come at a significant run-time cost.
Similarly, using a proposal network for sampling adds overhead at render time as multiple networks need to be evaluated sequentially.
To maximize rendering speed without model baking, we implement an explicit binary occupancy grid for efficiently skipping free space and minimizing the number of sample points for which hash grid features need to be queried and MLPs evaluated.
An example grid is shown in \cref{fig:4k-render}, along with the corresponding image results.
While occupancy grids are widely-adopted acceleration structures \cite{LiuGLCT2020,MuelleESK2022,SunSC2022,ChenXGYS2022}, we propose two novel extensions that help us prune more accurately.

\setlength{\columnsep}{5pt}
\begin{wrapfigure}[9]{r}{95pt}
\vspace{-1.1\baselineskip}
\includegraphics[width=\linewidth]{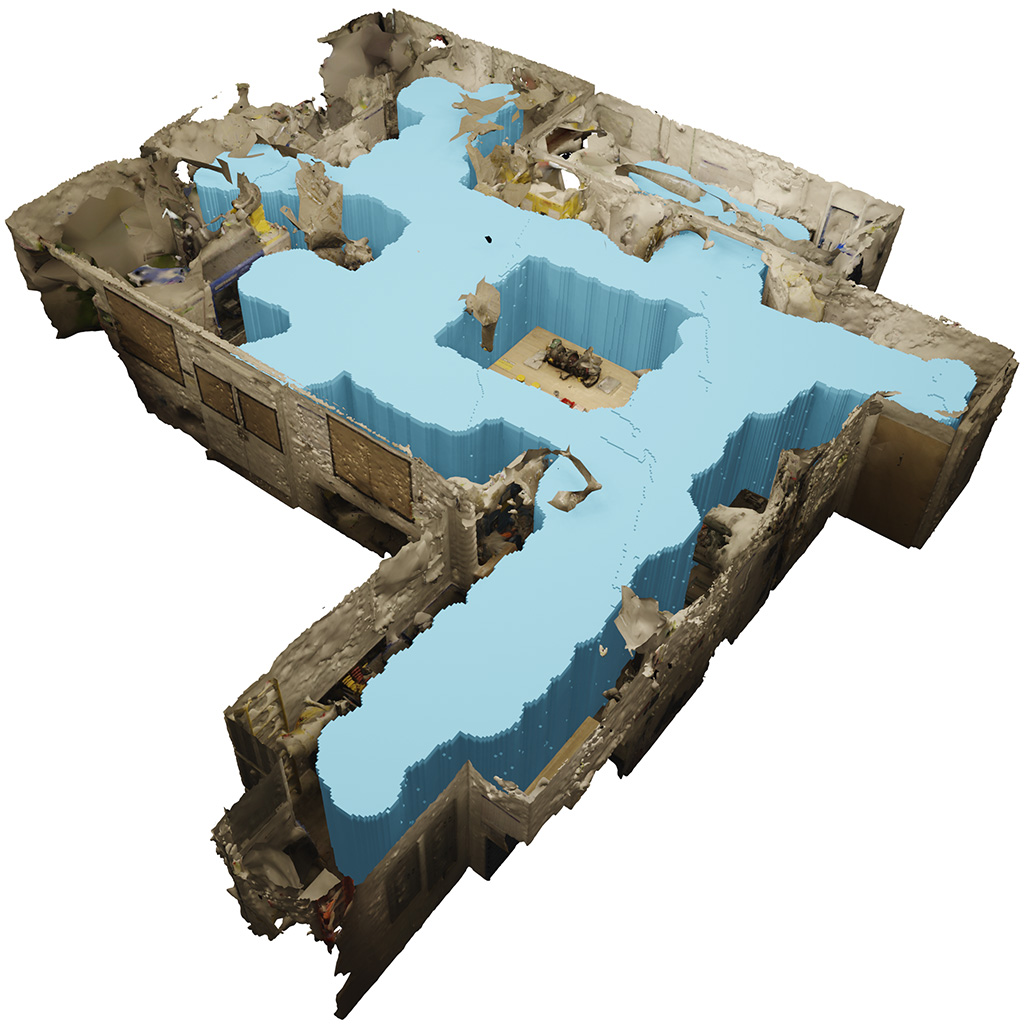}
\end{wrapfigure}
\subsubsection{Cylinder Pruning}

We initialize our binary occupancy grid based on the known rig capture positions to carve out as much free space in the scene as possible.
For this, we first approximate the geometry of our Eyeful Tower capture rig as a cylinder.
We then mark all occupancy grid voxels that are completely inside any such cylinder as free space.
This type of pruning has two key benefits:
(1) it prevents the model from cheating using floaters in front of cameras, which leads to more view-consistent models, and
(2) it speeds up the early stages of training and thus helps improve convergence speed.

\subsubsection{Joint History- and Grid-based Pruning}

We explore a more conservative pruning strategy that combines pruning based on training history with dense grid sampling.
History pruning keeps track of the maximum density observed for each voxel in the occupancy grid during the training process.
This only considers rays seen during training, so some parts of the scene may not be observed.
Grid-based pruning makes up for this by evaluating a dense cubic grid inside each voxel of the occupancy grid to estimate the maximum density for each voxel.
As the density depends on the step size used in training, we use a worst-case estimate for this, \ie, the minimum step size possible inside each voxel based on the ray from the closest camera.
For our datasets, we start the pruning process after 100K iterations, when a relatively clean scene geometry is obtained.
Every thousand iterations, we prune grid voxels for which both maximum densities fall below the current pruning threshold (which we anneal linearly from zero to $\alpha \!=\! 0.2$).
We start with a coarse occupancy grid of $128^3$ resolution, and upsample the occupancy grid at predefined iteration milestones to prune scenes more accurately over time.

\section{VR NeRF Rendering}
\label{sec:rendering}

Rendering a room-scale NeRF model in VR requires high resolution, high frame rates and low latency.
Our target is native rendering on a Meta Quest Pro VR headset, ideally dual 2K$\times$2K eyebuffers at 72\,FPS.
We approach this task with a combination of hardware, software, and model optimizations.
Specifically, we present a custom multi-GPU CUDA renderer with efficient in-register MLP evaluation and automatic work distribution, a compute-efficient LOD technique (see \cref{sec:lod}), and a 20-GPU workstation for peak VR performance.

MLP evaluation is the most computationally expensive portion of model inference, and thus a prime candidate for optimization.
Our MLP implementation is specialized for small iNGP-style networks by taking advantage of Nvidia's Tensor Cores and evaluating all layers within registers.
Inputs and outputs of the MLP are stored in shared memory while per-layer activations are stored in register-backed arrays, with outputs from one layer being shuffled in an architecture-dependent way to become the inputs to the next layer.
This limits memory traffic to just the input and output features, which are typically small (32 inputs, 16 bottleneck features, 3 output colors) compared to the hidden layers (64 nodes), and the network weights, which are shared across the kernel and typically cached.
This structure also allows the MLP evaluation to be interleaved with ray marching and hash grid sampling in a single kernel.
This enables the neural features to be passed to the networks without staging through global memory (which can suffer from capacity problems with a large number of samples per ray) or across multiple kernels (which would incur extra launch and synchronization overhead).

\begin{figure}
	\centering\setlength{\tabcolsep}{2pt}
	\begin{tabular}{ccc}
		\includegraphics[width=0.32\linewidth]{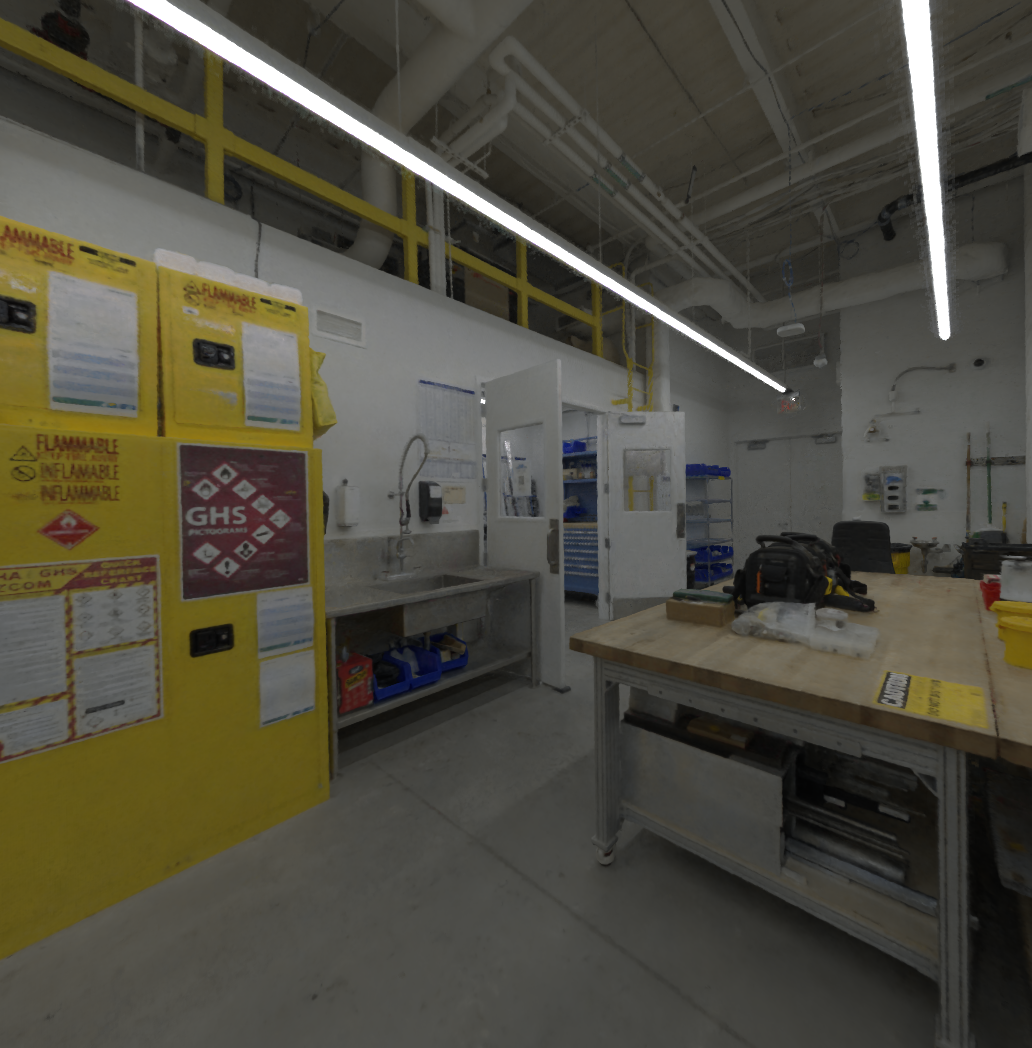} & 
		\includegraphics[width=0.32\linewidth]{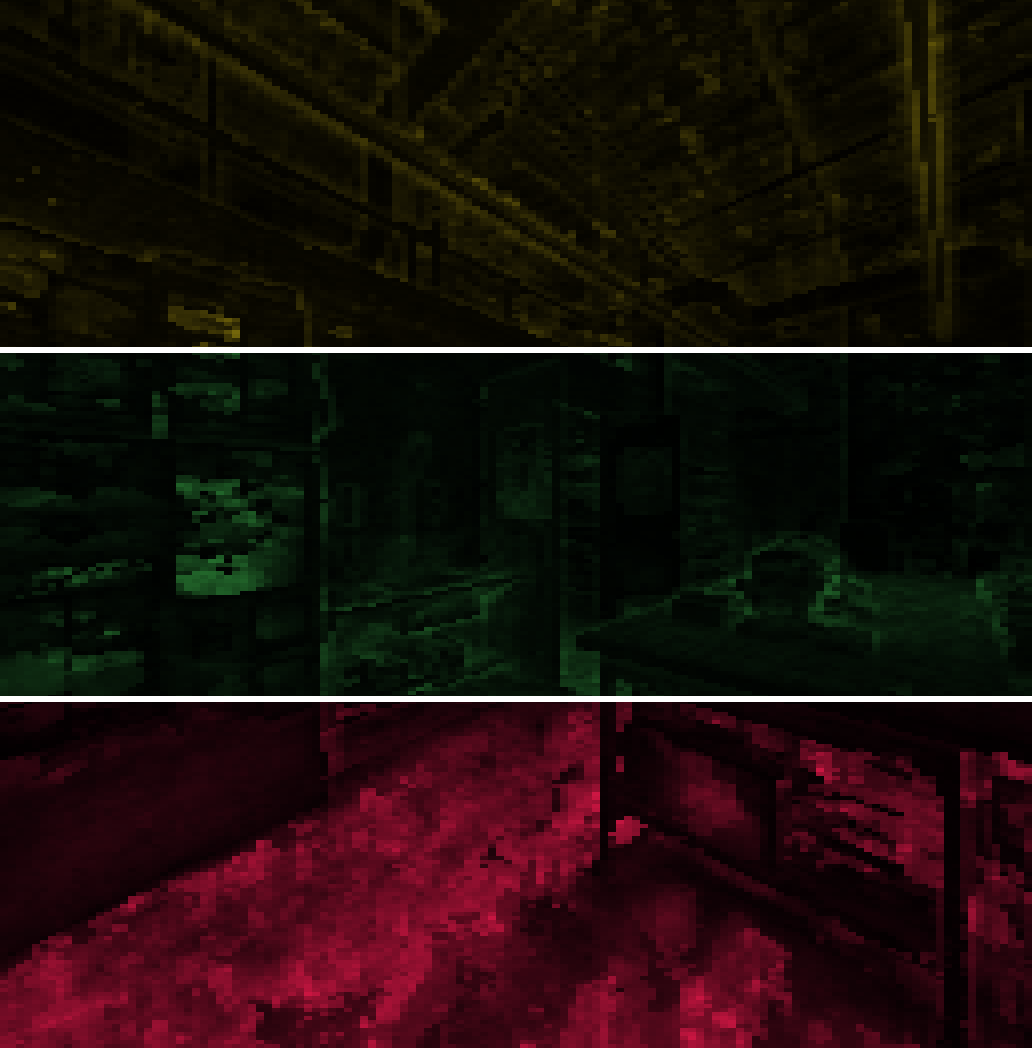} &
		\includegraphics[width=0.32\linewidth]{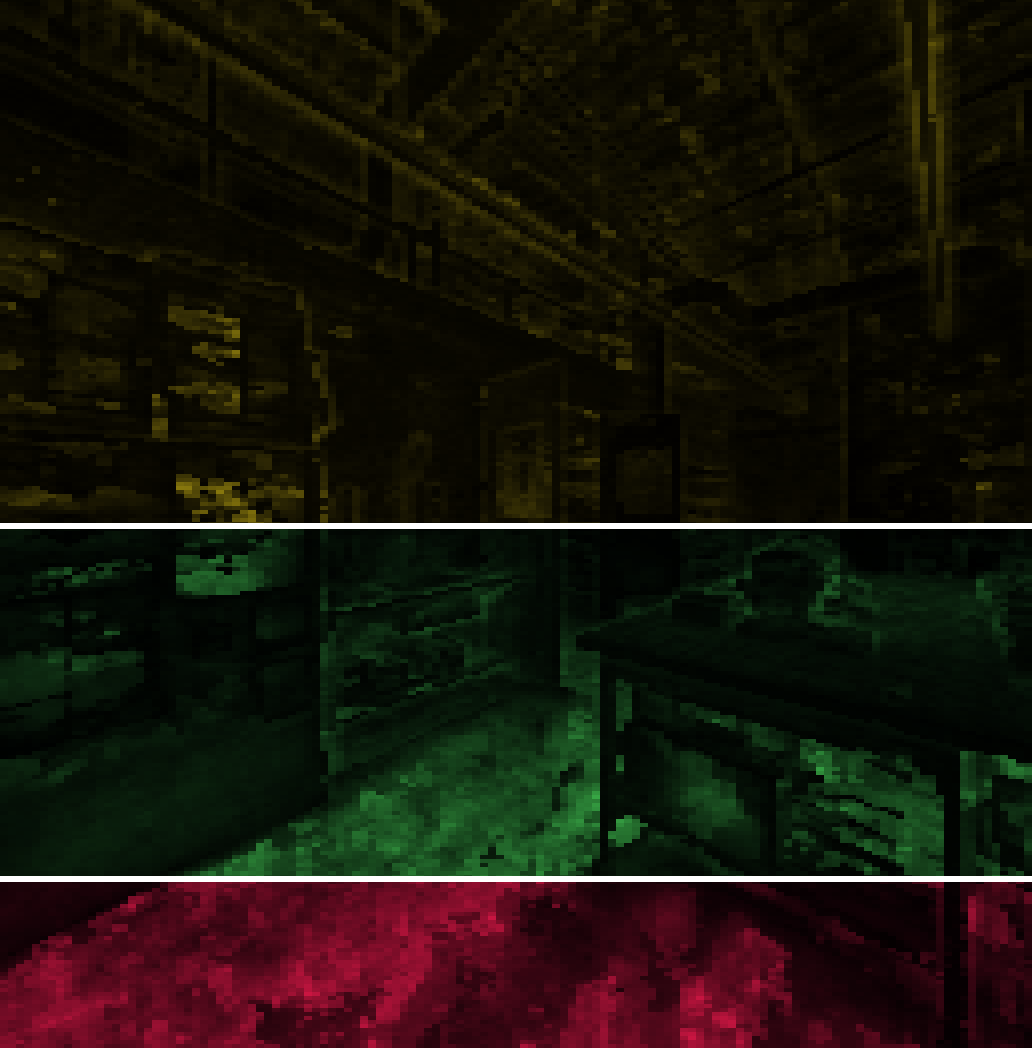} \\[-0.2em]
		\textsf{\scriptsize (a) Color image} &
		\textsf{\scriptsize (b) Static split: 49.3\,FPS} &
		\textsf{\scriptsize (c) Dynamic split: 73.6\,FPS}
	\end{tabular}
	\caption{\label{fig:work-distribution}%
		\textbf{(a)} In this example, we render a novel view using 3 GPUs.
		\textbf{(b)} A static split distributes work equally (indicated by colors; brightness is proportional to \#MLP evaluations).
		\textbf{(c)} Our dynamic work split achieves 49\% higher FPS.
	}
\end{figure}

We further adopt a dynamic work distribution strategy for improving the utilization of multiple GPUs compared to a static work split that would often be suboptimal as some rays take longer to compute than others due to differences in pruning in different parts of the scene, as well as GPU caching and overhead effects.
For every frame, we measure the throughput per GPU in rays per second, and assign contiguous rows to each GPU based on its ratio of the total throughput.
We use dampening for smoother convergence to an optimal distribution.
\Cref{fig:work-distribution} demonstrates a 49\% increase in FPS.
Each GPU stores a separate copy of each scene (\textasciitilde 700 MB VRAM).

We also built a custom 20-GPU rendering workstation to evaluate our walkable spaces at the highest possible fidelity in virtual reality.
This machine comprises a Dell R7515 server with an AMD Epyc 7313P CPU and 256\,GB of RAM, and is connected to 20 Nvidia A40 GPUs via a PCIe switching solution from Liqid Inc., all in a 24U server rack.
We detail our design considerations in the supplement.

\section{Results and Evaluation}
\label{sec:results}

For our room-scale scenes, we use hash grid configurations with $L\!=\!16$ levels of two features, with a base resolution of 128 and scaling factor of 1.4.
Following iNGP, we use a 1-hidden-layer density MLP and a 2-hidden-layer color MLP, both 64 neurons wide.
For each ray, we sample 1024 points using exponential distances for integration.
We use the Adam optimizer \cite{KingmB2015} with $\beta_1=0.9$, $\beta_2 = 0.99$, $\epsilon=10^{-15}$, and a batch size 12,800 rays (256 random rays from 50 random images) for all our experiments.
We use far-field contraction for the subset of unbounded scenes.
We use learning rate 0.01 for the hash grids and 0.005 for the remaining modules.
We discuss a series of additional techniques for per-scene quality improvements in the supplement.

For fair evaluation, we hold out a fixed camera from the training set, which has the same number of frames as all other cameras.
For ablation experiments, we show results trained for 110K iterations on 1K resolution Eyeful Tower datasets.
The demo videos are produced by models trained on 2K resolution images and longer than 200K iterations.
In the supplement, we include additional results on the Inria \cite{PhiliMGD2021} and mip-NeRF360 datasets \cite{BarroMVSH2022}, as well as ablations on pruning strategies.

\subsection{Comparative Evaluation}
\label{sec:comparisons}

To model HDR images, iNGP \cite{MuelleESK2022} suggests using an exponential color activation for linear RGB space.
RawNeRF \cite{MildeHMSB2022} further suggests using a weighted loss to prevent extremely bright areas from dominating.
\Cref{tab:avg_metric} and \cref{fig:qualitative} show the comparisons of our designed modules with iNGP baselines:
(1) the effectiveness of using PQ color space for HDR modeling, and (2) the use of the LOD feature grid.

We choose the baseline of using iNGP with linear color space with truncated exponential activation for the color network to avoid the issue of exploding values weighted by the predicted color value, as practiced by \citet{MildeHMSB2022}.
``iNGP with PQ color space'' reflects our modification of directly training in the PQ color space, and replaces the original exponential color activation with a sigmoid function.
``iNGP with PQ color space and LOD'' represents our core model of adopting mip-mapped grid features based on the estimated LOD level for each queried sample point on the ray.
We report the standard PSNR/SSIM/LPIPS metrics in tonemapped sRGB space, and additionally report versions of these metrics in PQ color space for better evaluation on extremely bright and dark areas.
More results and analysis with supporting plots and visualizations on each ablated module can be found in our supplement.

\begin{table}[t!]
    \caption{\label{tab:avg_metric}
        Quantitative comparison results on the Eyeful Tower test set.
        All results are trained on 1K resolution images for 110K iterations with 1024 samples per ray.
        We report the average PSNR/SSIM/LPIPS both in sRGB and PQ color spaces.
        The best results are \bb{highlighted}.
        See the supplemental document for the breakdown by individual dataset.
    }
    \sffamily
    \resizebox{\linewidth}{!}{%
    \begin{tabular}{l@{\hspace{12pt}}c@{\hspace{8pt}}c@{\hspace{8pt}}c@{\hspace{6pt}}c@{\hspace{2pt}}c@{\hspace{2pt}}c}
        \toprule
        Methods      & \scriptsize PSNR $\uparrow$ & \scriptsize SSIM $\uparrow$ & \scriptsize LPIPS $\downarrow$ & \tiny \scriptsize PQ-PSNR$\uparrow$ & \scriptsize PQ-SSIM$\uparrow$ & \scriptsize PQ-LPIPS$\downarrow$ \\
        \midrule
        iNGP (our implementation) &   31.93 & 0.918 & 0.183 & 37.39 & 0.957 & 0.133  \\
        with PQ color space  &   32.47 &     0.926  &     0.170  & 38.15 &     0.962  &     0.122  \\
        with PQ color space and LOD &  \bb{33.30} & \bb{0.930} & \bb{0.146} & \bb{38.95} & \bb{0.964} & \bb{0.108} \\
        \bottomrule
    \end{tabular}}
\end{table}

\paragraph{PQ color space}
\Cref{tab:avg_metric} shows that the PQ color space consistently outperforms the linear color space for all test scenes and all metrics.
We noticed that during training, color predictions using the exponential activation baseline continue to grow to excessively large values.
This poses an ambiguity for predicting correct density values and their derived weights, which are multiplied with the point color to obtain sample colors.
Directly modeling colors in linear RGB space poses additional challenges in regressing and interpolating accurate color values, especially when a large range of radiance is present.
As shown in the second example in \cref{fig:qualitative}, the base model fails to model the color on the checkerboard correctly.

\paragraph{Mip-mapped features}
The combination of LOD and PQ color space further improves the rendering quality and leads to cleaner geometries, as seen in \cref{tab:avg_metric} and \cref{fig:qualitative}.
This is critical for pruning and VR rendering, where a good geometry is desired.
\Cref{fig:qualitative} shows four scenarios where the mip-mapped features can help.
The first scene shows an annoying dark appearance baked into the rendered scene due to dynamic shadows from the rig in the training data.
These are modeled by the high-frequency levels and well addressed by the distance-aware features.
The second example shows both cleaner geometry and appearance in the ambiguous space near the whiteboard.
The third example shows how LOD helps eliminate aliasing for distant areas, especially when observing the scene from a wide angle.
The last example shows how LOD can also reveal more detail compared to non-mip-mapped features, not just a cleaner appearance.
This is expected as the features corresponding to high-resolution hash grids are specifically allocated to areas rich in fine details in the training views, which allows the model to automatically allocate more capacity for these parts.
Note that while the quantitative metrics are similar to the baselines, the visual improvements are easier to spot and critical to the high-fidelity rendering results that contribute to a pleasant VR experience.

\subsection{Performance Evaluation}
\label{sec:performance}

\begin{figure}

\pgfplotstableread[col sep=comma]{figures/performance-graphs-turtle-v4-fixed2.csv}{\TurtleData}

\begin{tikzpicture}[font=\small\sffamily]
\begin{groupplot}[
	group style={group name=performance, group size=2 by 2, vertical sep=0.9cm},
	height=4.0cm, width=4.7cm,
	cycle list/Paired,
	legend style={
		draw=none,
		at={(1.05,-2.2)},
		anchor=south,
		legend columns=3,
		/tikz/every even column/.append style={column sep=1.5em},
	},
	legend image post style={very thick},
	legend cell align={left},
]

\nextgroupplot[
	ymin=0, ymax=175, ytick={0,36,...,180}, ylabel={Frames per second}, axis y line=left, ymajorgrids=true,
	xtick={1,4,8,12,16,20}, xmax=21, axis x line=bottom,
	enlarge x limits=false,
	every axis plot post/.append style={mark=none},
	table/x=gpu-count,
	error bars/y dir=both, error bars/y explicit,
]

\draw[black,line width=1pt] (axis cs:1,72) -- (axis cs:25,72);
\draw[black,line width=1pt,dashed] (axis cs:1,36) -- (axis cs:25,36);

\addplot+ [thick] table [y=avg-fps|all|2x1K|eyeful15-city-of-bridges-1k, y error=std-fps|all|2x1K|eyeful15-city-of-bridges-1k] from \TurtleData; \addlegendentry{\textsc{office\_view1}};
\addplot+ [thick] table [y=avg-fps|all|2x1K|eyeful-roomsy-1k,            y error=std-fps|all|2x1K|eyeful-roomsy-1k]            from \TurtleData; \addlegendentry{\textsc{office2}};
\addplot+ [thick] table [y=avg-fps|all|2x1K|eyeful15-apartment1-1k,      y error=std-fps|all|2x1K|eyeful15-apartment1-1k]      from \TurtleData; \addlegendentry{\textsc{apartment}};
\addplot+ [thick] table [y=avg-fps|all|2x1K|eyeful15-seating-area-1k,    y error=std-fps|all|2x1K|eyeful15-seating-area-1k]    from \TurtleData; \addlegendentry{\textsc{seating\_area}};
\addplot+ [thick] table [y=avg-fps|all|2x1K|eyeful15-riverview-1k,       y error=std-fps|all|2x1K|eyeful15-riverview-1k]       from \TurtleData; \addlegendentry{\textsc{riverview}};
\addplot+ [thick] table [y=avg-fps|all|2x1K|eyeful-workshop-1k,          y error=std-fps|all|2x1K|eyeful-workshop-1k]          from \TurtleData; \addlegendentry{\textsc{workshop}};

\nextgroupplot[
	ymin=0, ymax=175, ytick={0,36,...,180},
	axis y line=left, ymajorgrids=true,
	xtick={1,4,8,12,16,20}, xmax=21, axis x line=bottom,
	enlarge x limits=false,
	every axis plot post/.append style={mark=none},
	table/x=gpu-count,
]

\draw[black,line width=1pt] (axis cs:1,72) -- (axis cs:25,72);
\draw[black,line width=1pt,dashed] (axis cs:1,36) -- (axis cs:25,36);

\addplot+ [thick] table [y=p99-fps|all|2x1K|eyeful15-city-of-bridges-1k, y error=std-fps|all|2x1K|eyeful15-city-of-bridges-1k] from \TurtleData;
\addplot+ [thick] table [y=p99-fps|all|2x1K|eyeful-roomsy-1k,            y error=std-fps|all|2x1K|eyeful-roomsy-1k]            from \TurtleData;
\addplot+ [thick] table [y=p99-fps|all|2x1K|eyeful15-apartment1-1k,      y error=std-fps|all|2x1K|eyeful15-apartment1-1k]      from \TurtleData;
\addplot+ [thick] table [y=p99-fps|all|2x1K|eyeful15-seating-area-1k,    y error=std-fps|all|2x1K|eyeful15-seating-area-1k]    from \TurtleData;
\addplot+ [thick] table [y=p99-fps|all|2x1K|eyeful15-riverview-1k,       y error=std-fps|all|2x1K|eyeful15-riverview-1k]       from \TurtleData;
\addplot+ [thick] table [y=p99-fps|all|2x1K|eyeful-workshop-1k,          y error=std-fps|all|2x1K|eyeful-workshop-1k]          from \TurtleData;

\nextgroupplot[
ymin=0, ymax=80, ytick={0,18,...,180}, ylabel={Frames per second}, axis y line=left, ymajorgrids=true,
xtick={1,4,8,12,16,20}, xmax=21, xlabel={Number of GPUs}, axis x line=bottom,
enlarge x limits=false,
every axis plot post/.append style={mark=none},
table/x=gpu-count,
error bars/y dir=both, error bars/y explicit,
]

\draw[black,line width=1pt] (axis cs:1,72) -- (axis cs:25,72);
\draw[black,line width=1pt,dashed] (axis cs:1,36) -- (axis cs:25,36);

\addplot+ [thick] table [y=avg-fps|all|2x2K|eyeful15-city-of-bridges-1k, y error=std-fps|all|2x2K|eyeful15-city-of-bridges-1k] from \TurtleData;
\addplot+ [thick] table [y=avg-fps|all|2x2K|eyeful-roomsy-1k,            y error=std-fps|all|2x2K|eyeful-roomsy-1k]            from \TurtleData;
\addplot+ [thick] table [y=avg-fps|all|2x2K|eyeful15-apartment1-1k,      y error=std-fps|all|2x2K|eyeful15-apartment1-1k]      from \TurtleData;
\addplot+ [thick] table [y=avg-fps|all|2x2K|eyeful15-seating-area-1k,    y error=std-fps|all|2x2K|eyeful15-seating-area-1k]    from \TurtleData;
\addplot+ [thick] table [y=avg-fps|all|2x2K|eyeful15-riverview-1k,       y error=std-fps|all|2x2K|eyeful15-riverview-1k]       from \TurtleData;
\addplot+ [thick] table [y=avg-fps|all|2x2K|eyeful-workshop-1k,          y error=std-fps|all|2x2K|eyeful-workshop-1k]          from \TurtleData;

\nextgroupplot[
ymin=0, ymax=80, ytick={0,18,...,180},
axis y line=left, ymajorgrids=true,
xtick={1,4,8,12,16,20}, xmax=21, xlabel={Number of GPUs}, axis x line=bottom,
enlarge x limits=false,
every axis plot post/.append style={mark=none},
table/x=gpu-count,
]

\draw[black,line width=1pt] (axis cs:1,72) -- (axis cs:25,72);
\draw[black,line width=1pt,dashed] (axis cs:1,36) -- (axis cs:25,36);

\addplot+ [thick] table [y=p99-fps|all|2x2K|eyeful15-city-of-bridges-1k, y error=std-fps|all|2x2K|eyeful15-city-of-bridges-1k] from \TurtleData;
\addplot+ [thick] table [y=p99-fps|all|2x2K|eyeful-roomsy-1k,            y error=std-fps|all|2x2K|eyeful-roomsy-1k]            from \TurtleData;
\addplot+ [thick] table [y=p99-fps|all|2x2K|eyeful15-apartment1-1k,      y error=std-fps|all|2x2K|eyeful15-apartment1-1k]      from \TurtleData;
\addplot+ [thick] table [y=p99-fps|all|2x2K|eyeful15-seating-area-1k,    y error=std-fps|all|2x2K|eyeful15-seating-area-1k]    from \TurtleData;
\addplot+ [thick] table [y=p99-fps|all|2x2K|eyeful15-riverview-1k,       y error=std-fps|all|2x2K|eyeful15-riverview-1k]       from \TurtleData;
\addplot+ [thick] table [y=p99-fps|all|2x2K|eyeful-workshop-1k,          y error=std-fps|all|2x2K|eyeful-workshop-1k]          from \TurtleData;

\end{groupplot}

\node[below = -2mm of performance c1r1.north] {mean$\pm$std at 2$\times$1K$\times$1K};
\node[below = -2mm of performance c2r1.north] {p99 at 2$\times$1K$\times$1K};
\node[below = -2mm of performance c1r2.north] {mean$\pm$std at 2$\times$2K$\times$2K};
\node[below = -2mm of performance c2r2.north] {p99 at 2$\times$2K$\times$2K};

\end{tikzpicture}
\caption{\label{fig:performance-graphs}%
    Runtime performance at half (\textbf{top}) and full (\textbf{bottom}) VR resolution (for a Meta Quest Pro) over a prerecorded camera trajectory.
    \textbf{Left:} Mean and standard deviation of frame rates.
    \textbf{Right:} The 99\textsuperscript{th} percentile frame time (expressed as FPS) is indicative of the worst-case frame rate.
}
\end{figure}
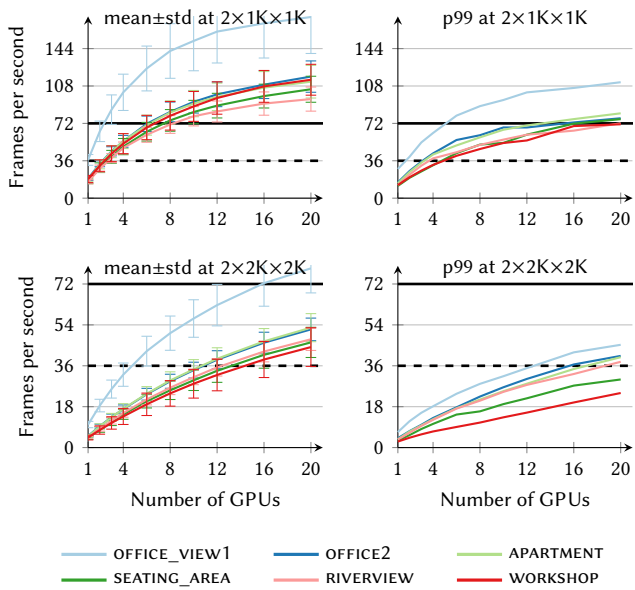

\begin{figure*}[p]
    \centering
    \includegraphics[width=0.99\linewidth]{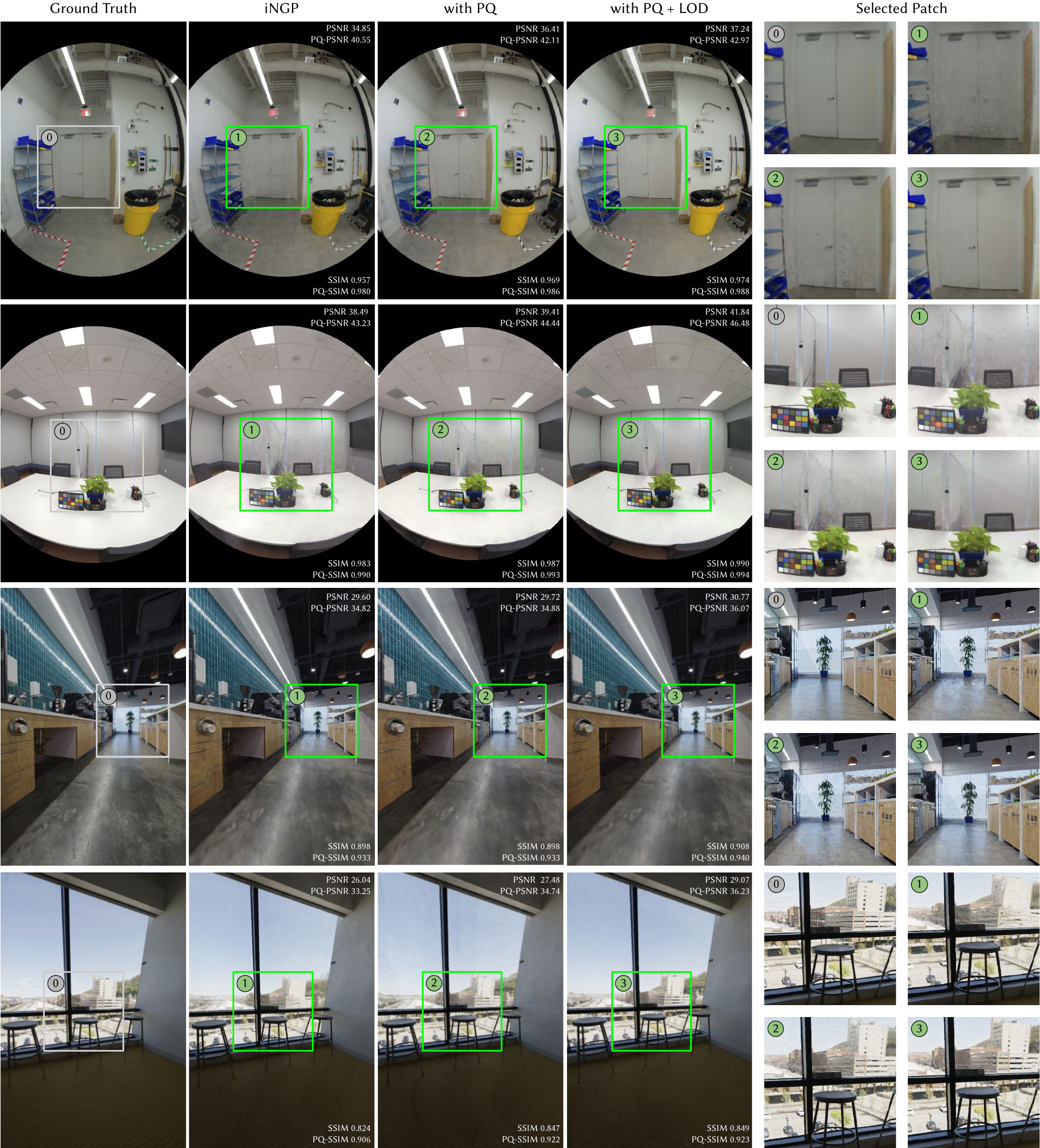}
    \caption{\label{fig:qualitative}%
        \textbf{Qualitative comparisons} between (1) iNGP, (2) iNGP with PQ color space, and (3) iNGP with PQ color space and LOD.
        The four selected examples show the improvements over the baselines by adding PQ color space and LOD in combination, which leads to cleaner appearance and geometry finer details.
        The use of PQ color space stablizes the learning of correct radiance values, while the inclusion of LOD helps to learn a cleaner appearance and geometry that is robust to distance-dependent appearance variations.
        By dynamically allocating model capacity to the sampled points based on the needed level of detail, it further reveals more details over the ablated counterparts.
        (Images are white-balanced and tonemapped for better visualization.)
    }
\end{figure*}

\begin{figure*}
	\centering
	\includegraphics[width=\linewidth]{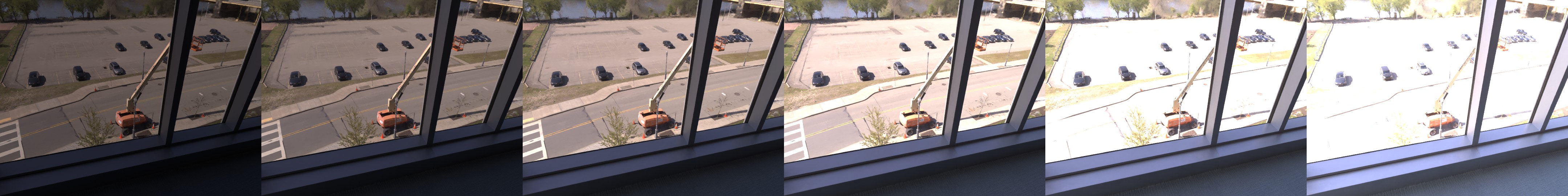}
	\caption{\label{fig:exposure-swipe}%
		\textbf{Sweep of exposure values}.
		For scenes with high dynamic range (\eg, bright outdoor views in the \textsc{riverview} scene), one can freely adjust the exposure setting at render time by manipulating the tonemapping from PQ color space to sRGB space.
	}
\end{figure*}

\begin{figure*}
    \centering
    \includegraphics[width=\linewidth]{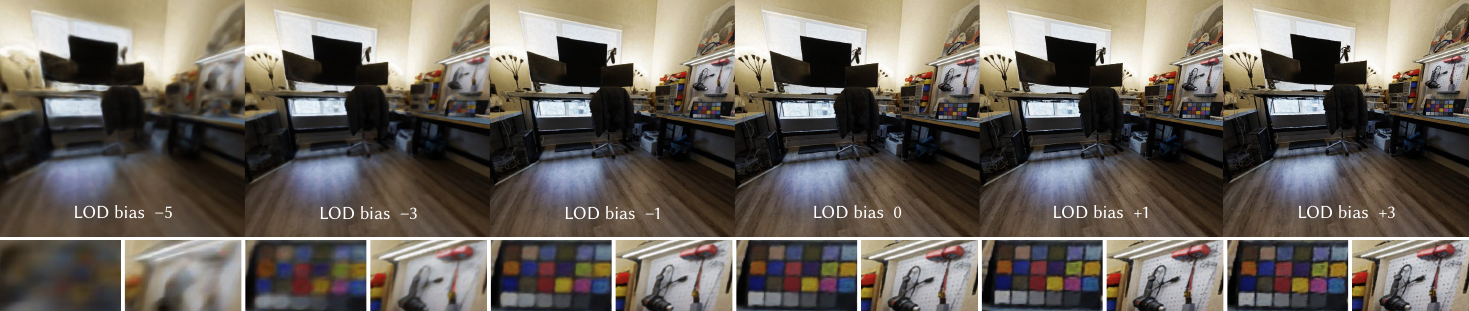}
    \caption{\label{fig:lod-swipe}%
        \textbf{Sweep of LOD bias}.
        We interpolate between a negative LOD bias of –5 and a positive LOD bias of +3 applied on top of the original estimated LOD value for each sample point on the \textsc{apartment} scene.
        A negative LOD bias blurs the rendering by masking out grid features representing high-frequency details, while a positive LOD bias helps reveal sharper details.
 }
\end{figure*}

\Cref{fig:performance-graphs} plots the rendering frame rate when using a varying number of A40 GPUs.
Native VR rendering for a Meta Quest Pro headset requires rendering two 2064$\times$2096 eyebuffers at 72\,FPS.
However, Asynchronous Spacewarp (ASW) \cite{BeeleHP2016} can help close this gap by reprojecting frames when they are rendered at least at half the native FPS, \ie, 36 (dashed line) instead of 72 (solid line).
But even ASW fails if rendering is slower than that critical threshold.
Thus, we found the `p99' metric (the 99\textsuperscript{th} percentile of FPS) to be a better proxy for the quality of VR experience than mean FPS --- as long as the application is typically (i.e. 99\%+ of the time) above 36 FPS, ASW can deliver a smooth experience.
Any slower, and the user may notice stuttering frames, and experience motion sickness.

An off-the-shelf 3-GPU workstation is sufficient for reliable half-resolution VR rendering of half the scenes at 36\,FPS (see \cref{fig:performance-graphs}, top right), which demonstrates the practicality of our method.
For maximum rendering speed and fidelity, we use our custom 20-GPU rendering workstation.
All scenes but one achieve a p99 of 72 FPS at half-resolution, and thus provide a smooth VR experience even without ASW.
At full resolution (top row in \cref{fig:performance-graphs}), 4 of the 6 scenes shown in \cref{fig:performance-graphs} (bottom right) exceed the critical ASW threshold of 36\,FPS for a smooth, high-fidelity VR experience.
Interestingly, halving the resolution only approximately doubles the FPS, even though only 1/4 as many pixels are being rendered.
This may indicate a substantial amount of per-frame overhead (e.g. due to kernel launches, the VR compositor, or OpenGL display pipeline) or insufficient parallelism available at lower resolutions.

\section{Discussion}
\label{sec:discussion}

\paragraph{Aggressive pruning}
Like most pruning approaches, we threshold density for determining if a voxel is occupied or not.
For bounded scenes with mostly solid surfaces, more aggressive pruning can be applied with a larger threshold (e.g., $\alpha \!=\! 0.3$) and finer grid resolution (e.g., $1024^3$), which results in significantly faster rendering (see `\textsc{office\_view1}' in \cref{fig:performance-graphs}).
However, aggressive pruning does not work well for complex real-world scenes, such as reflective surfaces, transparent objects or unbounded scenes.
This becomes particularly apparent in VR, where over-pruned areas show box-like artifacts that may not be easily seen in rendered 2D images or videos.

\paragraph{Distance-aware features}
Our level-of-detail feature weighting provides our model the flexibility to reproduce distance-dependent appearance such as varying level of detail, or rig shadows.
At the same time, we observed that this reduces our model's ability to extrapolate to unseen viewpoints or viewing distances, as feature vectors with unseen weighting may be used at render time.
In particular, density can vary depending on distance, which is undesirable.
We work around this by pruning as much free-space as possible, so that density cannot suddenly appear when moving through free-space.
\looseness-1

\section{Conclusion}
\label{sec:conclusion}

We presented VR-NeRF, the first holistic approach for capture, reconstruction and rendering of high-fidelity walkable spaces in virtual reality.
We made several key contributions across all stages of the pipeline to achieve the significantly higher resolution, frame rate and visual fidelity required for comfortable VR viewing of neural radiance fields.
We built a one-of-a-kind multi-camera rig that captures thousands of uniformly distributed HDR photos of a scene, integrated a novel perceptual color space for HDR model optimization, devised an efficient feature mip-mapping scheme for level-of-detail rendering, and implemented a multi-GPU renderer that achieves comfortable VR viewing on our demo machine.

\begin{acks}
We would like to thank Ada Lopaczynski, Autumn Trimble, Gadsden Merrill, Julia Buffalini, Kevyn McPhail, and Shukri Abdul Jalil for their exceptional technical support, and Alexandre Chapiro, Nathan Matsuda, Rafał Mantiuk and Yaser Sheikh for helpful discussions.
\end{acks}

\bibliographystyle{ACM-Reference-Format}
\bibliography{VR-NeRF}

\begin{figure*}
	\centering
	\includegraphics[width=\linewidth]{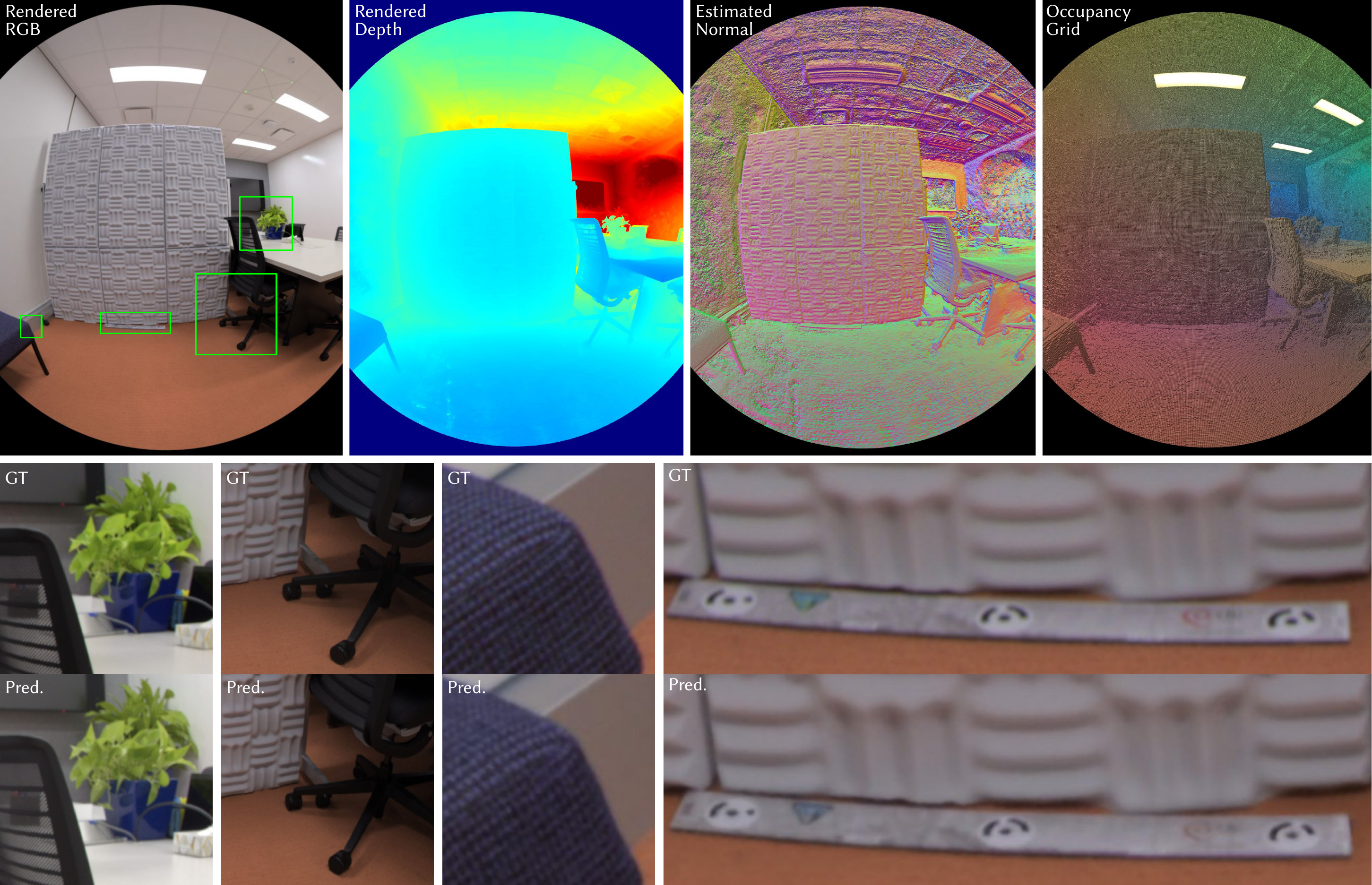}
	\caption{\label{fig:4k-render}%
		\textbf{\textsc{office2} results rendered at 4K resolution}, trained with 400K iterations.
		\textbf{Top row:} from left to right, we show (1)~rendered RGB image, (2)~estimated depth map, (3)~estimated normals, and (4)~the occupancy grid.
		\textbf{Bottom row:} The highlighted patches reveal sufficient fine details.}
\end{figure*}

\clearpage
\appendix
\section{Additional Capture Details}
\label{sec:capture-supp}

The system and procedure described in \mainorsup{\cref{sec:eyeful_tower}}{Section 3 of the main paper} is the third iteration of our capture stack.
The first version of the ``rig'' was a single handheld DSLR camera, but achieving the desired capture density quickly became tedious.
This was the primary motivation for the design of the second version of the rig, v1, which featured 9 cameras.
While a substantial improvement, the low camera count on v1 still required captures with the rig facing multiple directions.
Fortunately, v1 was designed with upgradability in mind, and we were quickly able to add an additional 13 cameras.
The result is the current configuration, v2, which again simplified the capture procedure and further increased our data quality.

\subsection{``Rig'' v0 — Handheld DSLR}

\paragraph{Capture hardware}
We began by performing handheld captures with a single Canon 1D X Mark II camera, which can take 20\,mega\-pixel photos.
This was paired with a Canon EF 8–15mm f/4L Fisheye USM lens, set to 8\,mm focal length to ensure the highest possible field of view, which minimized the number of images required to achieve high viewing direction coverage.
A cellphone camera was initially considered, but ultimately rejected due to the lack of interchangeable lens and insufficient field of view on the existing lenses.
We also experimented with using a tripod for additional stability during capture, but found it too cumbersome to continuously reposition it and thus removed it.

\paragraph{Capture procedure}
Each capture began by picking a reasonable ISO, shutter speed, and f-stop which would be fixed for the scene -- typically somewhere around ISO 1000, 1/40 seconds, and $f/4$.
The lens was set to its widest setting, at 8\,mm.
We then walked multiple loops around the scene, taking an image at small steps, typically about 30\,cm, along each one.
The camera would be held level horizontally, and its height would be increased each loop, starting at approximately knee height and increasing in 20–50\,cm intervals until the camera was above the head height of the person performing the capture.
This typically resulted in approximately 100--200 photos being captured for a single scene in 1--2 hours.

\subsection{Eyeful Tower v1 — 9 fisheye cameras}
The captures we performed with the single handheld camera were enough to get us started, but also clearly had some limitations.
The biggest, and most obvious, was the difficulty and time required to ensure sufficient scene coverage.
We also observed over- and under-exposure of different parts of our scenes, such as when looking through windows (at the sun) or at shadows.
We endeavored to address both of these issues with a rig that featured multiple rigidly mounted cameras, at differing heights, which could capture simultaneously.
Cameras at different heights would allow captures similar to the handheld setup with only a single pass of the scene, rather than one per height as before.
The rigid mounting would also enable an exposure bracket to be taken, enabling HDR images to be generated.

\paragraph{Camera, lens, and exposure}
The construction of a multi-camera rig gave us the opportunity to take a closer look at our camera and lens selection.
After carefully considering options for both professional and machine vision cameras, we chose the Sony $\alpha$1 mirrorless interchangeable lens cameras for striking a good balance between resolution (50\,megapixels), dynamic range (14 stops), ease-of-use (available analog triggers and Ethernet socket), as well as form factor (smaller than our previous Canon DSLR camera).
We continued to use the same lens as before, via a Metabones EF-to-E-mount adapter, but this time zoomed to 12\,mm to fill more of the camera's sensor, as shown in \cref{fig:v1_sample}.
We use ISO 500, the camera's higher native ISO, for minimal imaging noise, and set the aperture to $f/8$ with a focus distance of 1\,m for a large depth of field.
The cameras are configured to take a 9-image exposure bracket, 1 stop apart (–4 EV to +4 EV), as shown in \cref{fig:exposure_stops}.
The center exposure value is adjusted per-scene and is typically between 1/200 and 1/60 of a second.
RAW and JPEG images are stored redundantly on the two SD cards in each camera.

\begin{figure}
	\centering
	\includegraphics[width=\linewidth]{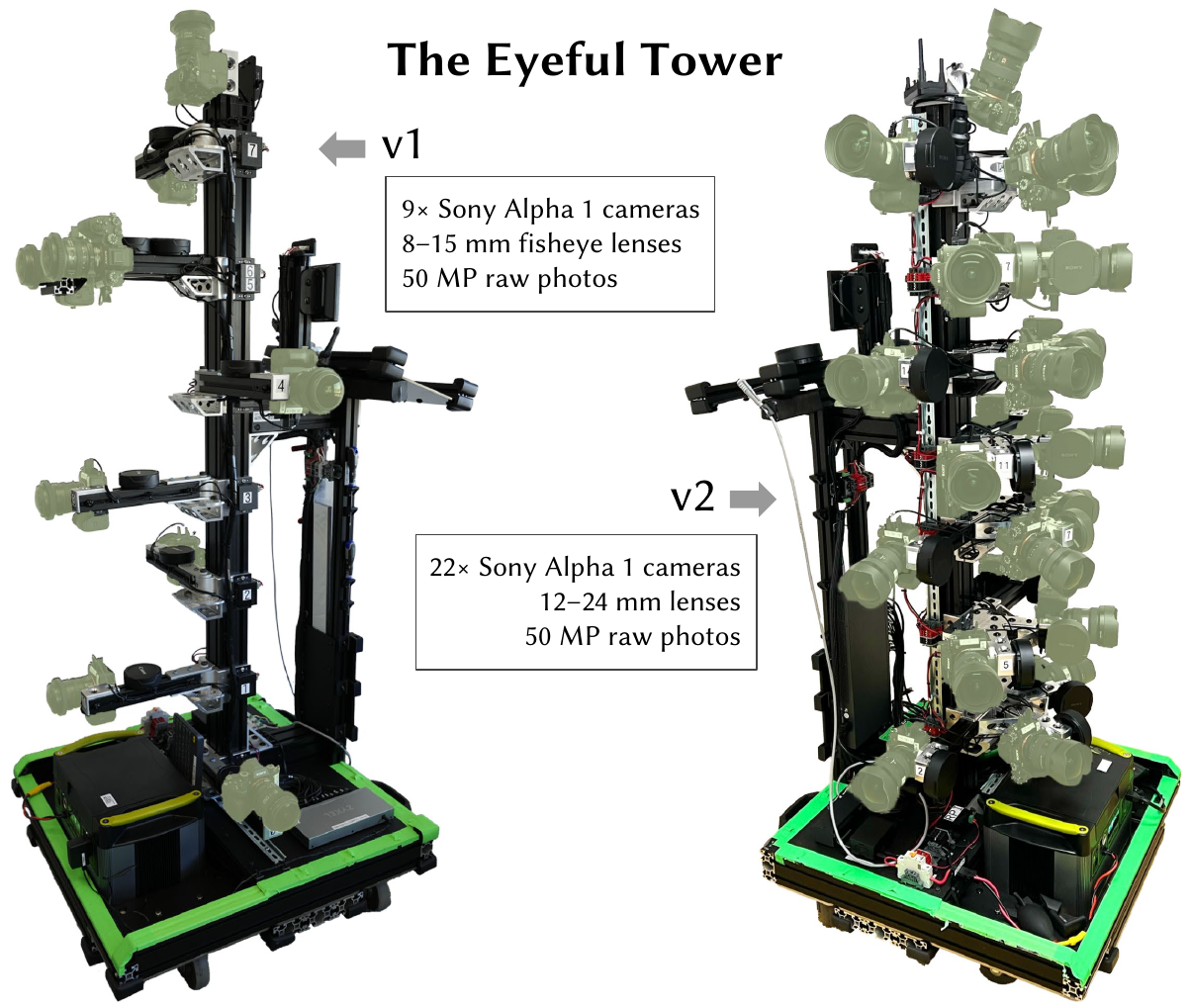}
	\caption{\label{fig:eyeful_tower}%
            The two versions of the Eyeful Tower camera rig.
	}
\end{figure}

\begin{figure*}
	\centering
	\includegraphics[width=0.8\linewidth]{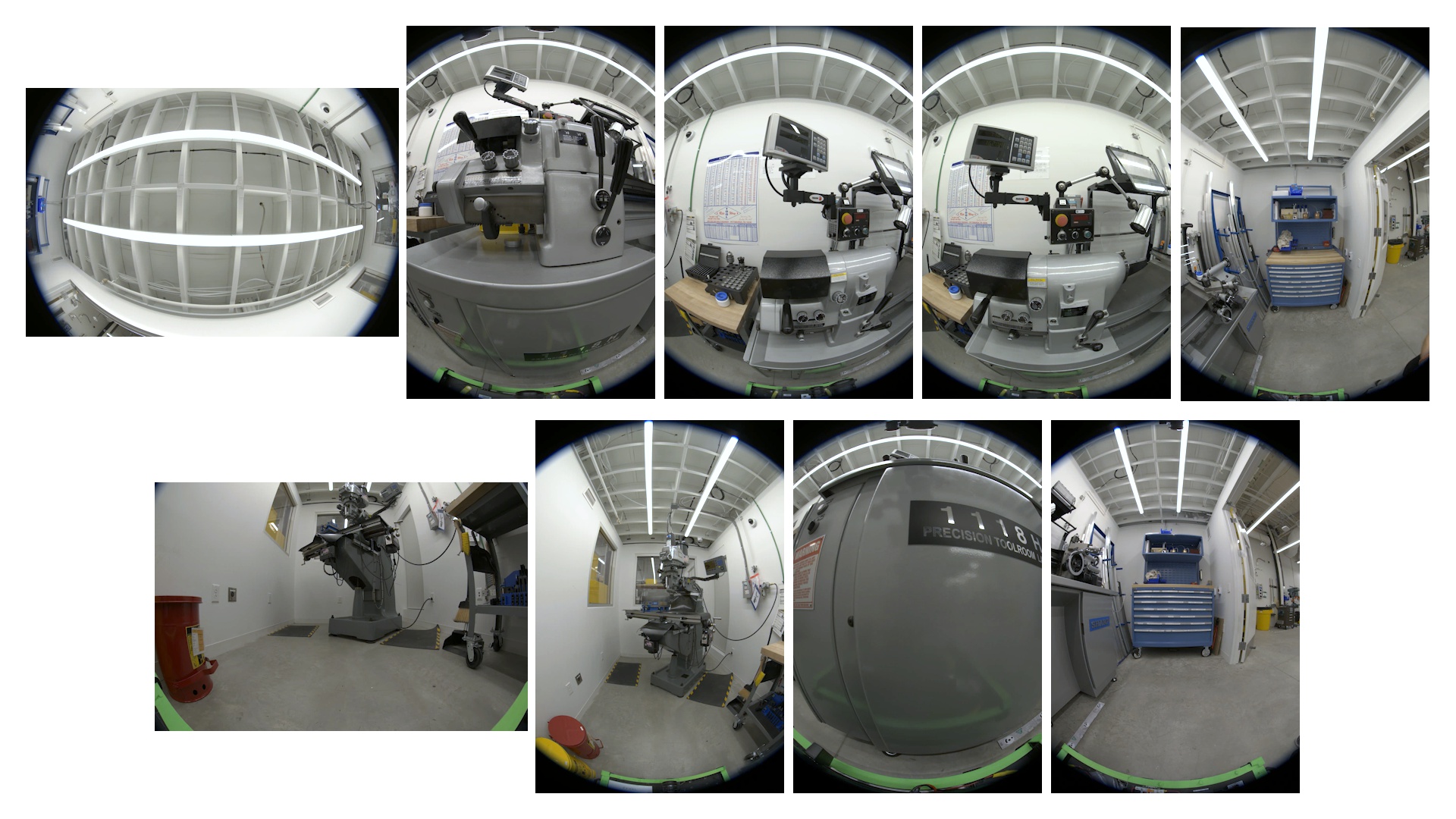}
	\caption{\label{fig:v1_sample}%
		Example frames from the `\textsc{workshop}' dataset captured using the 9 cameras in Eyeful Tower v1.
	}
\end{figure*}

\begin{figure*}
	\centering
	\includegraphics[width=0.8\linewidth]{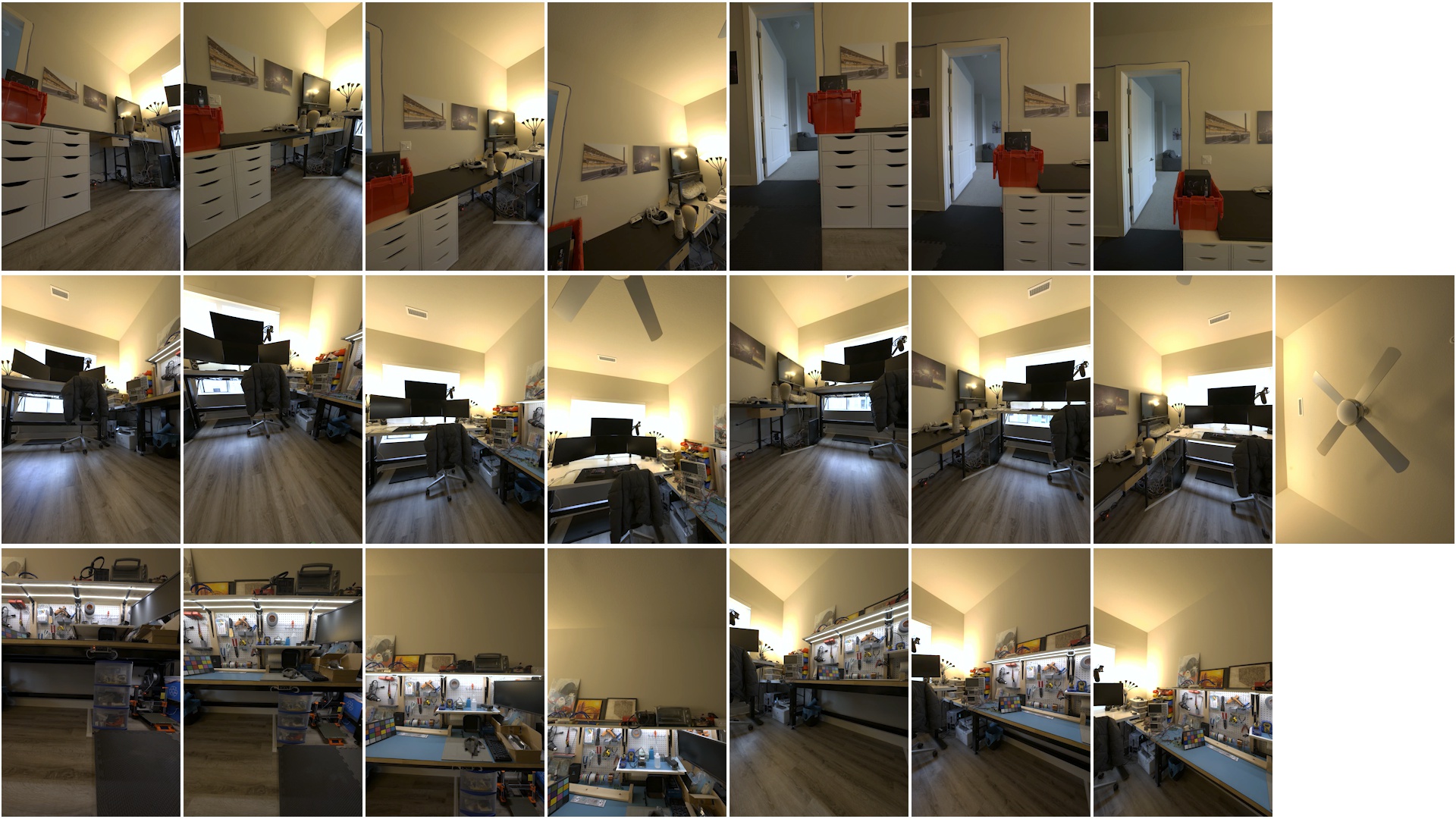}
	\caption{\label{fig:v2_sample}%
        Example frames from the `\textsc{apartment}' dataset captured using the 22 cameras in Eyeful Tower v2.
	}
\end{figure*}

\begin{figure*}
	\centering
	\includegraphics[width=\linewidth]{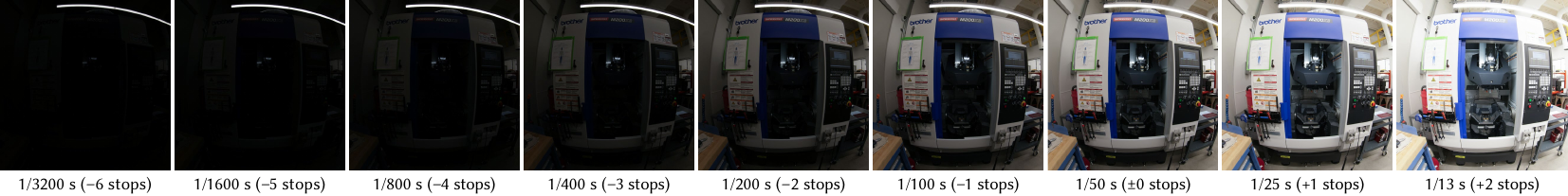}
	\caption{\label{fig:exposure_stops}%
		To capture the high dynamic range (HDR) of the real world, we take nine exposures at increasing shutter speeds, and merge these photos into a single HDR image.
		We can therefore reproduce the full dynamic range of input exposures, including the brightest and darkest regions.
	}
\end{figure*}

\paragraph{Mechanical design and camera placement}
We designed the capture rig using 80/20 extruded aluminium around an 80\,$\times$\,80\,cm base with a 1.8\,m vertical pole, allowing for substantial adjustability and expandability.
The pole held 7 camera brackets, each capable of supporting a single camera, whose height could be adjusted, and whose direction could be adjusted within 180 degrees horizontally.
We positioned the cameras left, forward, and right in an alternating fashion, as shown on the left side of \cref{fig:eyeful_tower}, attempting to maximize scene coverage from a single rig position while allowing sufficient space for the operator to not be visible in camera images when standing behind the rig.
One forward-facing bracket, at roughly eye height, was specially modified to support a second camera that would be held out for validation.
A final camera was added at the top for ceiling coverage, which was otherwise not present, for a total of 9 cameras on v1.

\paragraph{Electrical design}
At the base of the rig is a 1.5\,kWh Li-ion battery, capable of supplying regulated 12\,V DC and 120\,V AC power via an integrated inverter.
The 12\,V bus is used to provide power to all cameras and a Raspberry Pi 4 via a 12\,V to USB-PD adapter.
The cameras are powered by internal Li-ion batteries, but are continually trickle charged via USB.
The 120\,V AC output powers a 24-port 1-Gigabit Ethernet switch that connects all the cameras to the Raspberry Pi.
The Pi runs custom capture software to enable formatting of all camera SD cards, camera parameter updates, and simultaneous bracket triggering via the Sony Camera Remote SDK.
The switch also offers a 10GbE SFP+ port which is used to offload data from the cameras to a PC for downstream processing.
The battery is able to power the entire system for 6 to 8 hours of use, and can be recharged via normal 120\,V wall power overnight.

\begin{figure}[b]
	\centering\setlength{\tabcolsep}{2pt}
	\begin{tabular}{cc}
		\includegraphics[height=0.42\linewidth]{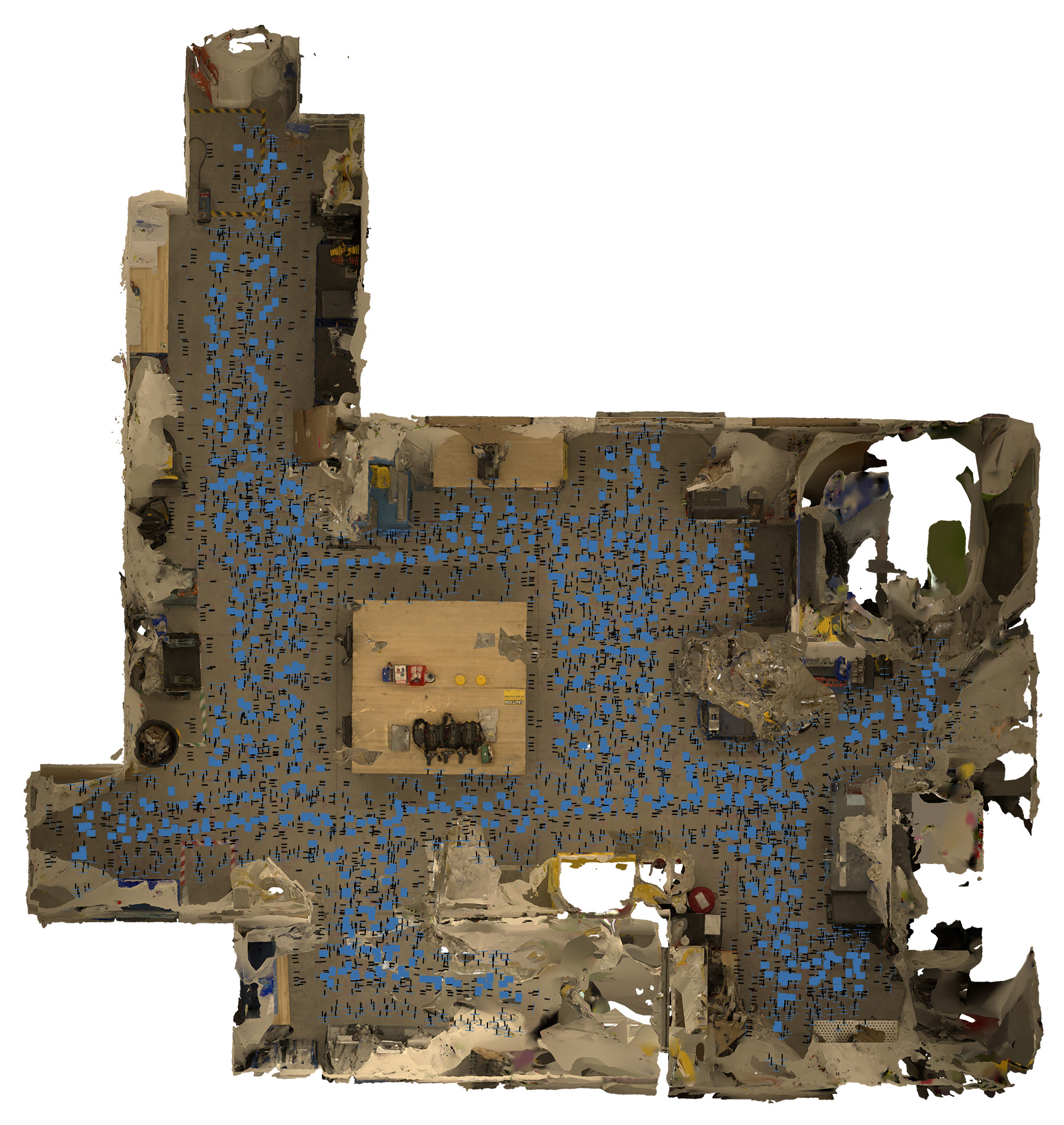} & 
		\includegraphics[height=0.42\linewidth]{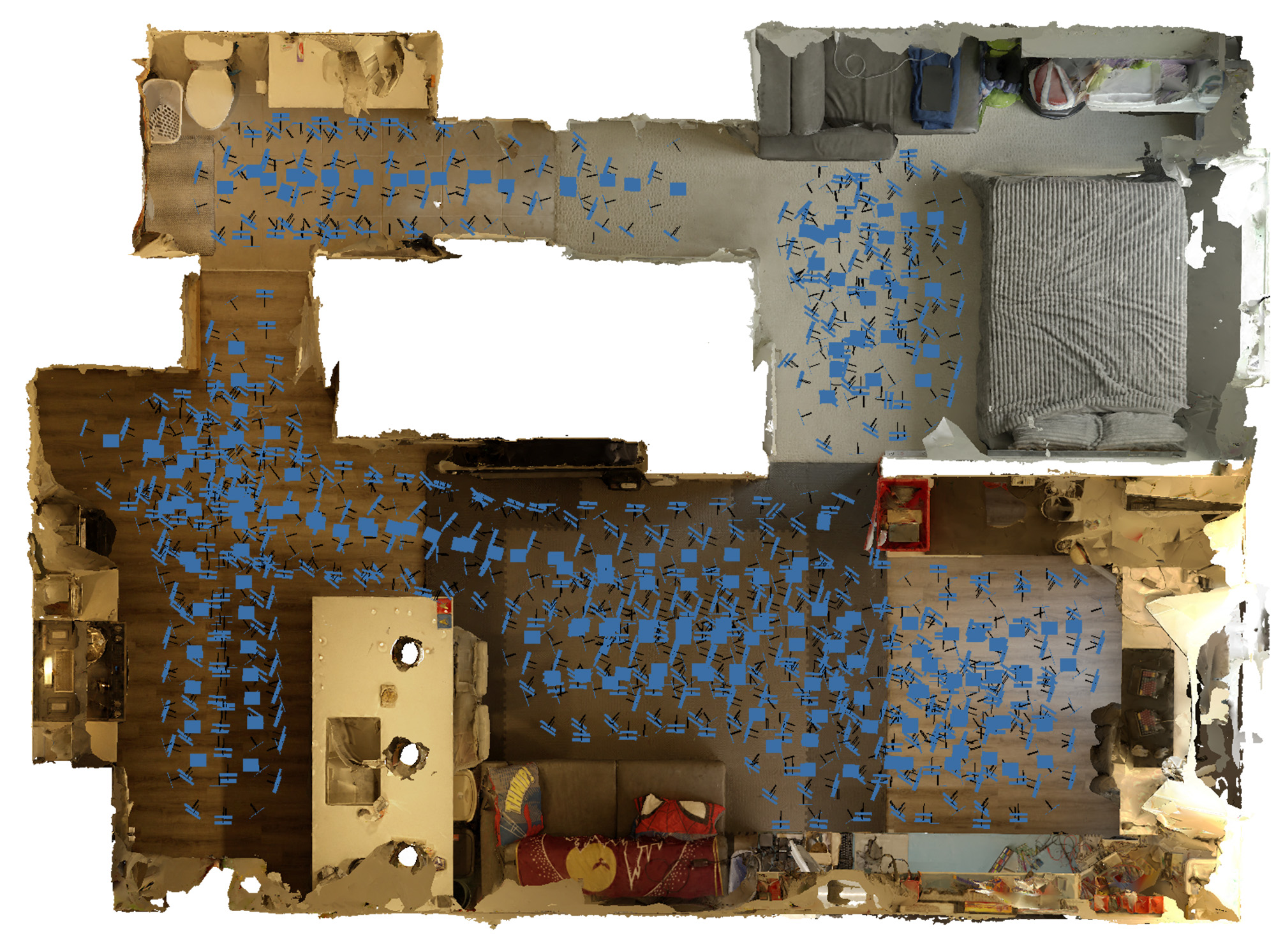}
	\end{tabular}
	\caption{\label{fig:scene_overview}%
        Visualized camera positions for the \textsc{workshop} (left) and \textsc{apartment} (right) scenes.
	}
\end{figure}

\paragraph{Capture procedure}
Cameras at multiple heights and viewing directions simplifies the coverage problem during capture from 6D to 3D, as we now only need to tile the floor with the rig facing a few directions.
The general capture strategy is described in \mainorsup{\cref{sec:capture-procedure}}{Section 3.2.2 of the main paper}, with example camera positions shown in \cref{fig:scene_overview}.
For this initial version of the rig, however, we needed to capture in four separate orientations (facing forward, backward, left, and right, versus just forward and backward as described above), in order to ensure 360 horizontal degrees of coverage for each rig position at each height.

\subsection{Eyeful Tower v2 — 22 rectilinear lenses}

The additional cameras of Eyeful Tower v1 offered a substantial usability and data quality improvement over the single handheld camera, but the sparse positioning still necessitated four passes of the scene.
The current version of the rig, v2, attempts to address this by adding yet more cameras, increasing the total count to 22, as shown on the right in \cref{fig:eyeful_tower}.
The lenses have also been replaced with rectilinear Sony FE 12–24mm F/2.8 GM lenses, selected for their higher sharpness (2–3× MTF50) and lower chromatic aberration (less than half, in pixels).
The camera and lens parameters are kept the same as before --- 12\,mm zoom, ISO 500, $f/8$ aperture, and 1\,m focus distance.
An example set of captured images is shown in \cref{fig:v2_sample}.
The additional cameras more than compensate for the slightly lower per-camera FOV, and enable us to capture a scene in two passes (rig facing forward and backward) rather than the four that v1 required.

\section{Model Implementation Details}

\begin{figure}[b]
	\centering
	\includegraphics[width=\linewidth]{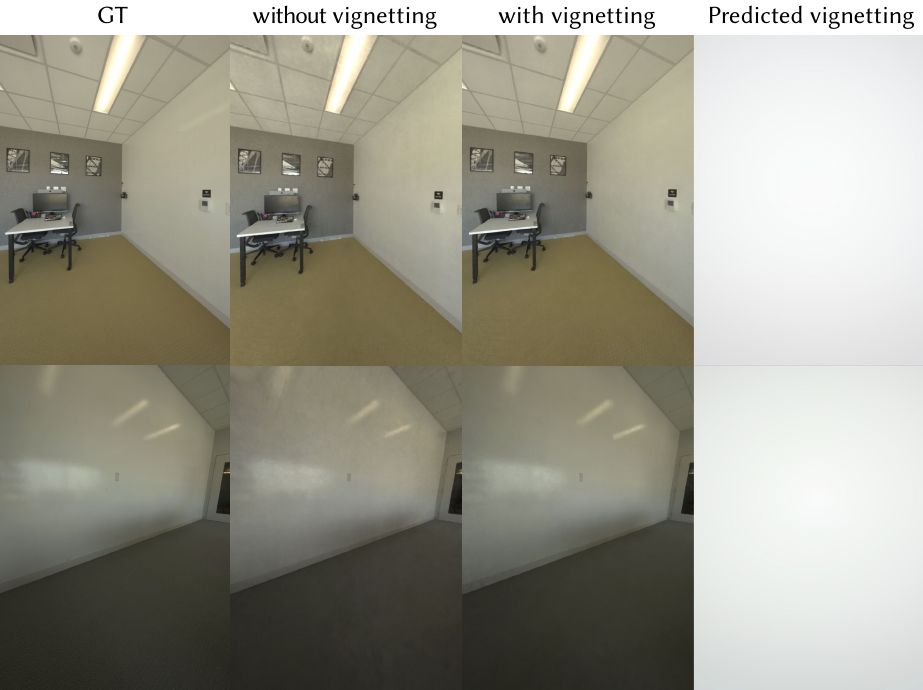} \\
    \vspace{1mm}
    \includegraphics[width=\linewidth]{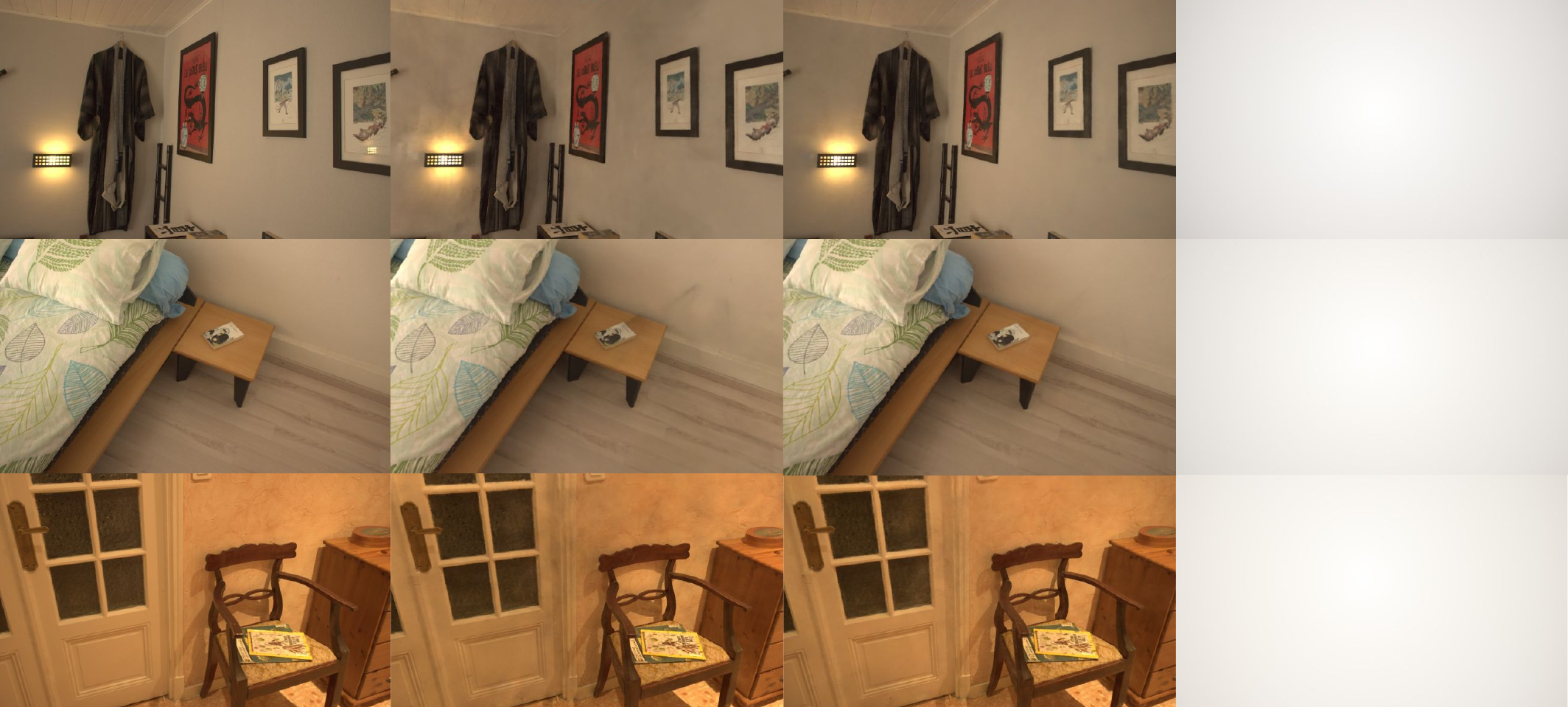}
	\caption{\label{fig:vignetting}%
		\textbf{Ablation on vignetting effects.}
		Without explicitly modeling vignetting, the model has difficulty in explaining away the inconsistent appearance brought by different camera lenses.
		We found that this effect is easier to observe in textureless areas, such as walls and floors in \textsc{office\_view1} (first row) and \textsc{office\_view2} (second row), with noticeable dark floaters in front of cameras.
		The vignetting effect is found more strongly in the Inria datasets \cite{PhiliMGD2021}, as shown in the bottom figure.
	}
\end{figure}

\subsection{Vignetting and Lens Distortion}

Vignetting effects are commonly present in wide-angle lenses, such as the fisheye and wide-angle lenses used in our capture system (see \cref{sec:capture-supp}).
We parameterize the vignetting effect for each camera using the \citeauthor{KangW2000} model \citeyearpar{KangW2000} with $I' \approx (1-\alpha r) I$, where we make $\alpha$ and the principal point $(c_x,c_y)$ used for computing $r$ learnable parameters for each camera sensor.
This allows the model to fit to the radial falloffs.
Without modeling vignetting \cite{Lyu2010} explicitly, the model is likely to overfit on training views by casting unwanted black floats everywhere in the air to accounting for the brightness decrease towards the edge of each image frame.
\Cref{fig:vignetting} shows that modeling vignetting explicitly is crucial for certain subset of data.
In scenarios where the test camera has unlearned vignetting parameters, one can optionally optimize the vignetting model for the test camera lens before inference with all the other parameters fixed.
The optimization can be done quickly in hundreds or thousands iterations.
Note that we do not optimise for lens distortions \cite{XianBSL2023} in our current implementation, which is left as future work to further improve pixel-wise alignment and accuracy.

\subsection{Loss Design}

\paragraph{Image Loss}
With PQ color space ranged in [0,1], we can directly use $L_1/L_2$ for reconstruction loss metric without biasing towards certain color range.
We prioritize $L_1$ loss in our experiments, as it is generally regarded as a more robust loss for outliers and provide sparser solutions compared to $L_2$ loss.
We also experimentally found that $L_1$ leads to sharper details compared to $L_2$ loss.

\begin{figure}[b]
    \centering
    \includegraphics[width=\linewidth]{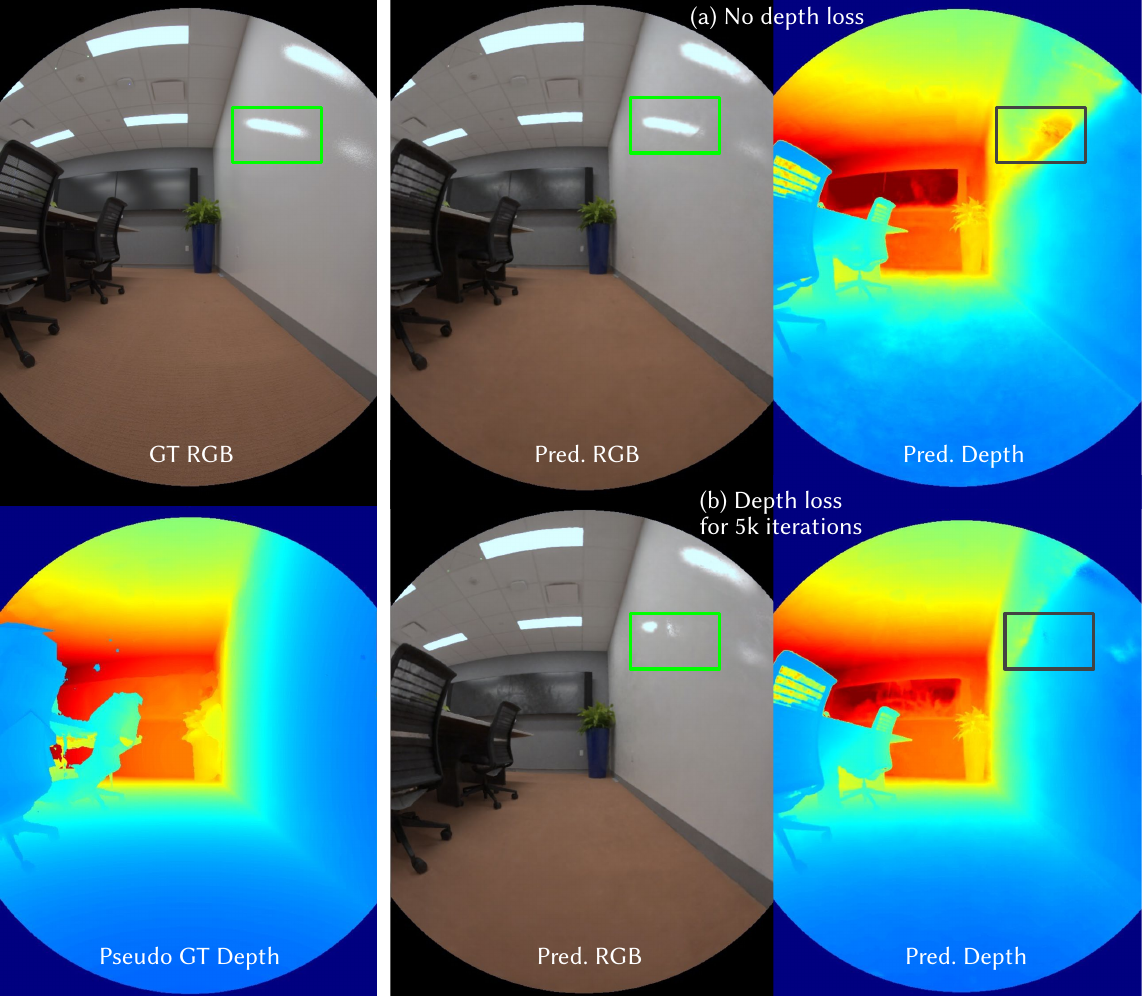}
    \caption{\label{fig:depth-loss}%
        \textbf{Applying depth loss.}
        The bottom-left image shows the pseudo depth guidance obtained from Metashape's mesh reconstruction.
        Experimentally, we found that the depth loss converges extremely fast in few hundreds or thousands iterations.
        The first row shows the results without depth supervision, while the result shown in the second row is supervised by the depth loss for the first 5K iterations.
        The depth supervision prevents model from cheating reflections with samples placed behind the walls, leading to the inability to recover the highlights.
    }
\end{figure}

\paragraph{Depth Regularizations}
NeRF reconstruction can be challenging for textureless areas, such as a flat white wall or a featureless floors.
Furthermore, the challenging reflections and shadows which caused the abrupt change in the brightness can easily lead to incorrect geometry where the view-dependent effects fail to capture the variances.
To address this issue, some approaches use additional geometrical information \cite{YuPNSG2022}, such as depth map guidances.
As the monocular depth predicted by off-the-shelf models can only be used in relative scale, we instead resort to the reconstructed mesh from Metashape during the pose estimation stage and project it to the training views as pseudo ground truth depth map for supervision, which allows us to perform direct comparison in absolute scale.
To avoid the misguidance from unreliable depth map, we use depth loss only for the early stage of training, where the incorrect geometry can get refined with image reconstruction loss only.
\Cref{fig:depth-loss} shows the effects of using depth loss guidance in early training stages.

\begin{figure}[b]
	\centering
	\includegraphics[width=\linewidth]{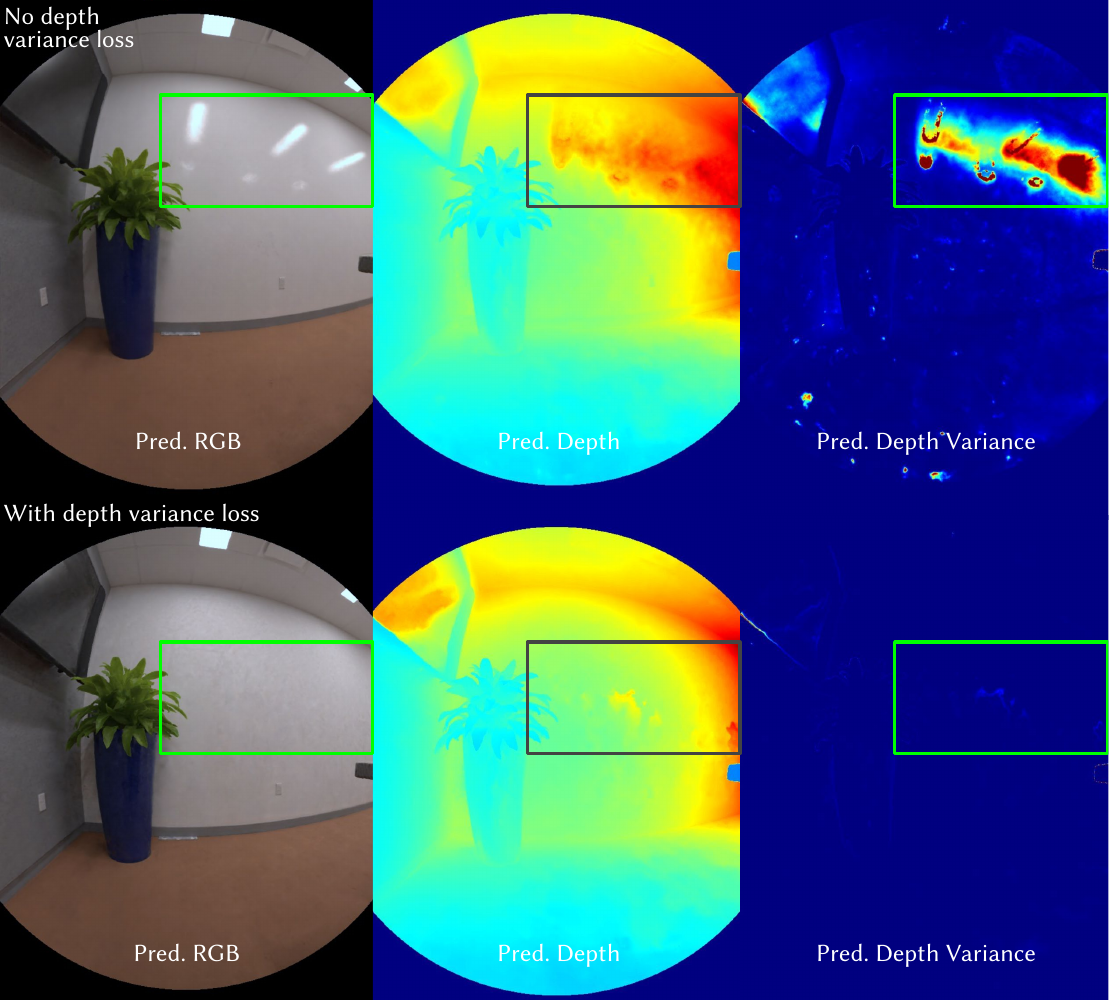}
	\caption{\label{fig:depth-var-loss}%
	    \textbf{Applying depth variance loss.}
	    Depth variance loss can be applied without ground-truth depth supervision.
	    The depth variance map on the top right shows a clear correlation between these reflection regions and the depth variance statistics, which indicates its potential to suppress cheating of appearance changes via wrong geometry.
	}
\end{figure}

\paragraph{Distortion Loss}
We adopt a simplified version of distortion loss \cite{BarroMVSH2022} that encourages the sparseness of the sample weights along the ray without considering the compactness of sample intervals from the proposal network.
In practice, we additionally consider the depth variance loss applied on the inner world (regions unaffected by space warping) that encourages the weights to be concentrated around the estimated depth.
The depth variance is calculated among samples cast by a single pixel.
\Cref{fig:depth-var-loss} shows the effects of applying depth variance loss that learns flat wall without reflections.
This trick is suggested to use during the separate pruning stage, where the cleaner geometry is used for obtaining reliable occupancy grid only.
This serves as an additional regularization for challenging scenarios such as the highly reflective surfaces, planes with detailed textures, where the depth variance could be relative large leading to incorrect geometry.
We optionally add an ``empty around camera'' loss by placing random samples in the unit sphere around the cameras to avoid near-plane ambiguity, similar to the occlusion loss in FreeNeRF \cite{YangPW2023}.

\paragraph{Other Regularizations}
\citet{BarroMVSH2023} recently proposed to apply a weight-decay loss on the multi-level hash grid features to encourage a normal distribution of learned grid features.
We found that this loss can regularize the learning of grid features, leading to more complete geometry and flat surfaces, yet at the cost of lower convergence speed and occasional detail loss.

\begin{table*}
	\caption{\label{tab:full_ablate}\label{tab:full_results}%
        Quantitative comparison results on the Eyeful Tower test set.
        All results are trained on 1K images for 110K iterations with 1024 samples per ray.
        We report PSNR/SSIM/LPIPS both in sRGB and PQ color spaces.
        The best results are \bb{highlighted}.
    }
	\centering
	\def\sqm{m\textsuperscript{2}}%
	\renewcommand*{\arraystretch}{1.15}%
    \setlength{\tabcolsep}{4pt}%
    \sffamily
	\resizebox{\linewidth}{!}{%
        \begin{tabular}{l@{\hspace{12pt}}c@{\hspace{12pt}}c@{\hspace{8pt}}c@{\hspace{8pt}}c@{\hspace{4pt}}c@{\hspace{1pt}}c@{\hspace{1pt}}c@{\hspace{16pt}}c@{\hspace{8pt}}c@{\hspace{8pt}}c@{\hspace{4pt}}c@{\hspace{1pt}}c@{\hspace{1pt}}c@{\hspace{16pt}}c@{\hspace{8pt}}c@{\hspace{8pt}}c@{\hspace{4pt}}c@{\hspace{1pt}}c@{\hspace{1pt}}c}
            \toprule
                                    &    Far-    & \multicolumn{6}{c}{\hspace{-12pt}iNGP (our implementation)}
                                                 & \multicolumn{6}{c}{\hspace{-16pt}iNGP with PQ color space}
                                                 & \multicolumn{6}{c}{\hspace{-12pt}iNGP with PQ color space and LOD} \\
            Scene                   &   field    & \scriptsize PSNR $\uparrow$ & \scriptsize SSIM $\uparrow$ & \scriptsize LPIPS $\downarrow$ & \tiny \scriptsize PQ-PSNR$\uparrow$ & \scriptsize PQ-SSIM$\uparrow$ & \scriptsize PQ-LPIPS$\downarrow$
                                                 & \scriptsize PSNR $\uparrow$ & \scriptsize SSIM $\uparrow$ & \scriptsize LPIPS $\downarrow$ & \tiny \scriptsize PQ-PSNR$\uparrow$ & \scriptsize PQ-SSIM$\uparrow$ & \scriptsize PQ-LPIPS$\downarrow$
                                                 & \scriptsize PSNR $\uparrow$ & \scriptsize SSIM $\uparrow$ & \scriptsize LPIPS $\downarrow$ & \tiny \scriptsize PQ-PSNR$\uparrow$ & \scriptsize PQ-SSIM$\uparrow$ & \scriptsize PQ-LPIPS$\downarrow$ \\
			\midrule
			\textsc{apartment}      & \checkmark & 30.68 & 0.903 & 0.226 & 35.16 & 0.942 & 0.203 & 31.06 &     0.910  &     0.208  & 35.52 &     0.946  &     0.186  & \bb{32.36} & \bb{0.915} & \bb{0.190} & \bb{36.61} & \bb{0.948} & \bb{0.173} \\
			\textsc{kitchen}        & \checkmark & 31.33 & 0.925 & 0.216 & 36.43 & 0.954 & 0.166 & 31.50 &     0.928  &     0.210  & 36.67 &     0.956  &     0.161  & \bb{32.41} & \bb{0.932} & \bb{0.184} & \bb{37.53} & \bb{0.958} & \bb{0.146} \\
			\textsc{office1a}       &            & 35.71 & 0.972 & 0.095 & 41.75 & 0.988 & 0.055 & 36.20 &     0.974  &     0.091  & 42.43 &     0.989  &     0.052  & \bb{36.82} & \bb{0.976} & \bb{0.082} & \bb{43.05} & \bb{0.990} & \bb{0.048} \\
			\textsc{office1b}       &            & 27.58 & 0.880 & 0.402 & 33.38 & 0.952 & 0.248 & 28.44 &     0.895  &     0.361  & 34.59 &     0.960  &     0.216  & \bb{29.97} & \bb{0.914} & \bb{0.255} & \bb{36.16} & \bb{0.966} & \bb{0.162} \\
			\textsc{office2}        &            & 39.84 & 0.983 & 0.053 & 44.46 & 0.986 & 0.031 & 40.26 & \bb{0.992} &     0.046  & 45.12 &     0.993  &     0.026  & \bb{40.71} &     0.987  & \bb{0.037} & \bb{45.53} & \bb{0.994} & \bb{0.023} \\
			\textsc{office\_view1}  & \checkmark & 29.75 & 0.890 & 0.264 & 35.14 & 0.942 & 0.192 & 30.20 &     0.897  &     0.253  & 35.63 &     0.947  &     0.181  & \bb{31.80} & \bb{0.901} & \bb{0.223} & \bb{37.09} & \bb{0.948} & \bb{0.167} \\
			\textsc{office\_view2}  & \checkmark & 27.19 & 0.858 & 0.259 & 33.14 & 0.921 & 0.201 & 27.64 &     0.865  &     0.259  & 33.73 &     0.926  &     0.198  & \bb{28.08} & \bb{0.868} & \bb{0.230} & \bb{34.09} & \bb{0.927} & \bb{0.181} \\
			\textsc{riverview}      & \checkmark & 27.17 & 0.864 & 0.181 & 34.09 & 0.934 & 0.135 & 27.83 & \bb{0.869} & \bb{0.179} & 34.77 & \bb{0.938} & \bb{0.134} & \bb{28.47} &     0.863  &     0.181  & \bb{35.51} &     0.935  &     0.135  \\
			\textsc{seating\_area}  & \checkmark & 36.85 & 0.968 & 0.070 & 41.93 & 0.983 & 0.050 & 37.16 &     0.970  &     0.069  & 42.26 &     0.984  &     0.049  & \bb{37.86} & \bb{0.974} & \bb{0.053} & \bb{42.92} & \bb{0.987} & \bb{0.035} \\
			\textsc{table}          & \checkmark & 34.28 & 0.952 & 0.091 & 40.37 & 0.974 & 0.067 & 34.52 &     0.954  &     0.087  & 40.78 &     0.975  &     0.064  & \bb{35.40} & \bb{0.962} & \bb{0.066} & \bb{41.59} & \bb{0.980} & \bb{0.046} \\
			\textsc{workshop}       &            & 30.86 & 0.907 & 0.155 & 36.54 & 0.949 & 0.115 & 32.34 &     0.937  &     0.102  & 38.12 &     0.965  &     0.074  & \bb{32.40} & \bb{0.940} & \bb{0.101} & \bb{38.31} & \bb{0.967} & \bb{0.073} \\
			\midrule
			Average                 &            & 31.93 & 0.918 & 0.183 & 37.39 & 0.957 & 0.133 & 32.47 &     0.926  &     0.170  & 38.15 &     0.962  &     0.122  & \bb{33.30} & \bb{0.930} & \bb{0.146} & \bb{38.95} & \bb{0.964} & \bb{0.108}  \\
            \bottomrule
		\end{tabular}%
	}
\end{table*}

\begin{figure}[b]
	\centering
	\includegraphics[width=\linewidth]{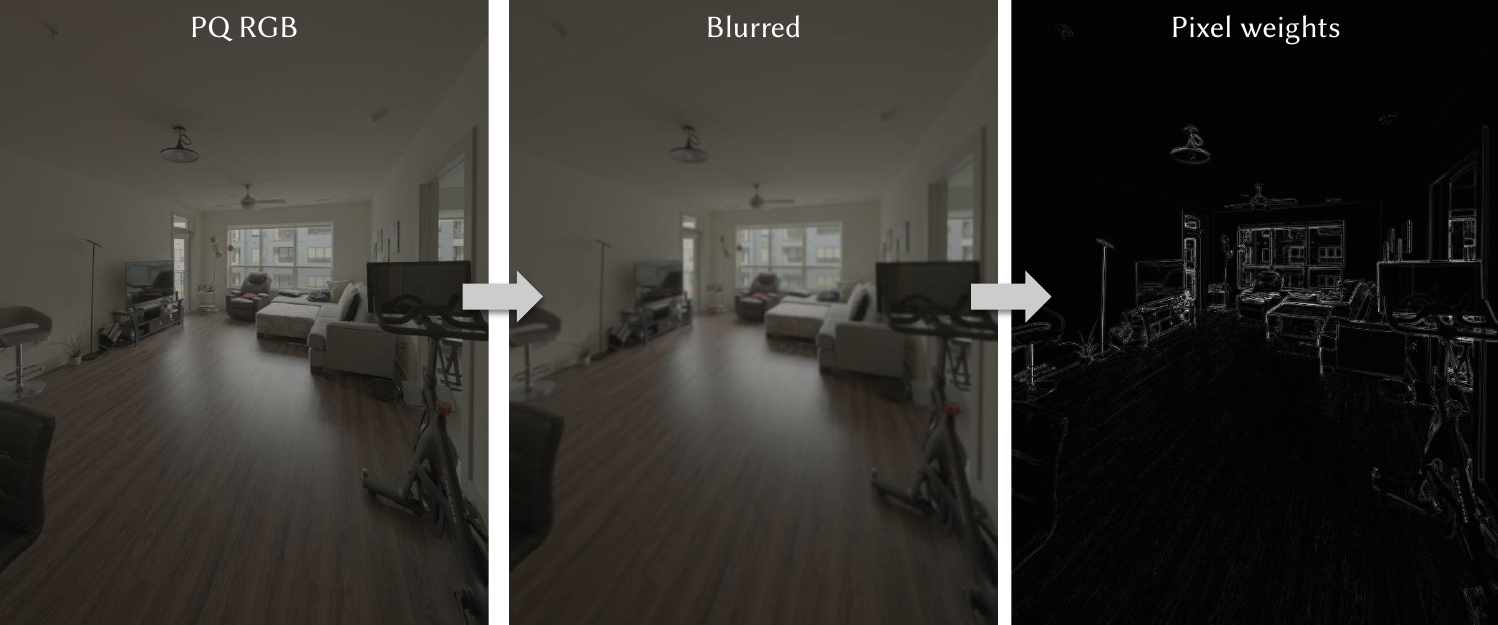}
	\caption{\label{fig:hardray}%
		\textbf{Derivation of pixel weights.}
		We borrow the idea of Laplacian pyramid and derive the importance weight for each pixel to guide sampling.
	}
\end{figure}

\begin{figure}[t]
	\centering
    \includegraphics[width=\linewidth]{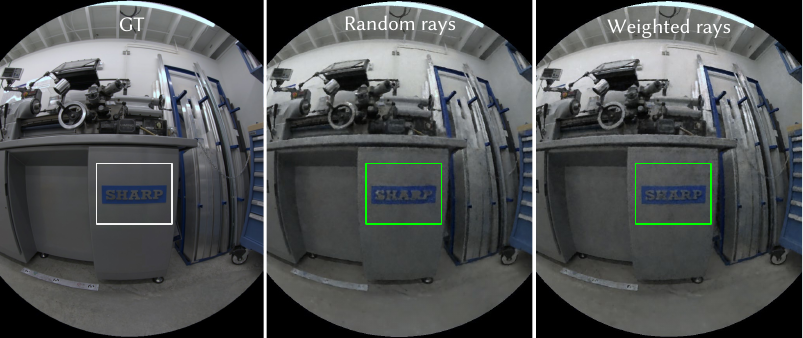}
	\caption{\label{fig:hardray2}%
		\textbf{Weighted samples.}
		Effects of revealing finer details and geometry in early training stages (\eg, 2K iterations) with weighted sampling.
		The sign of `SHARP' in the scene is quickly revealed with weighted sampling.
	}
\end{figure}

\begin{figure}[b]
	\centering
	\includegraphics[width=\linewidth]{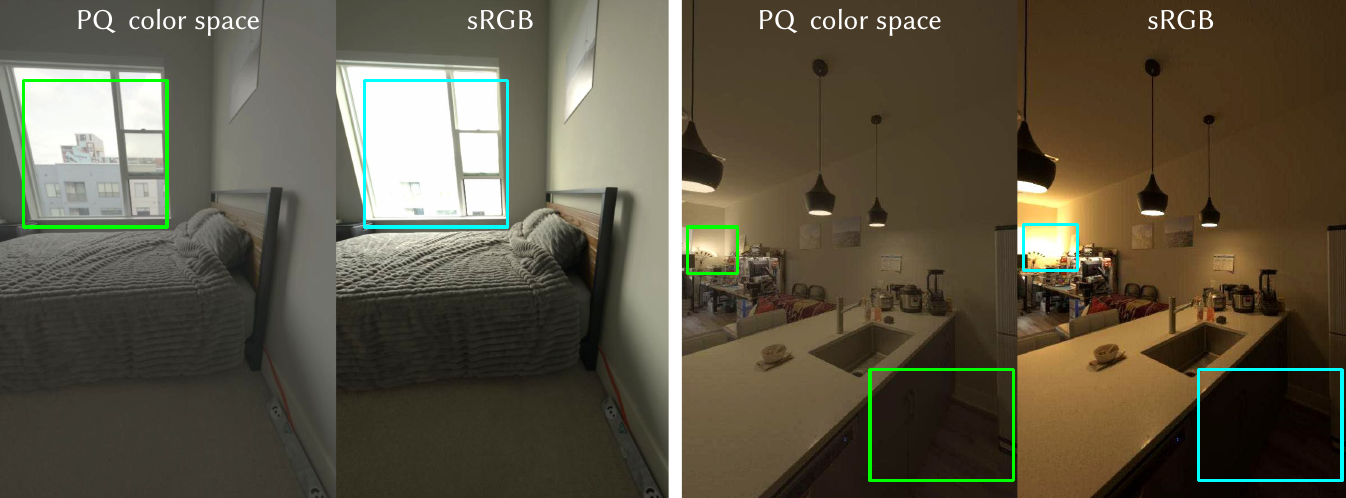}
	\caption{\label{fig:pq-space}%
		\textbf{Illustration of PQ color space.}
		The highlighted regions represent extreme bright/dark areas which are properly handled by the PQ conversion.
	}
\end{figure}

\subsection{Weighted Sampling}

Given the large number of rays used to be supervised during training, it is generally impractical to revisit each pixel multiple times with limited training time.
The training of NeRF is notorious for long training time in order to bring finer details, which usually appear in the later stage of training due to the spectral bias of neural networks \cite{RahamBADLHBC2019}.
We also observed from the computed error map that the errors usually dominate in high-frequency details, while areas with relative uniform geometry and appearance have significant lower errors.
One can either assign higher loss weights to these pixels representing high-frequency details or sampling them more often during the training.
Practically, we compute the Laplacian pyramid for each image during data preparation stage and use them as the indicator for important pixel samples.
Instead of using the directly obtained continuous value for pixel weights, we threshold the importance samples by referencing to the 75\% percentile of all pixels over the image.
See \cref{fig:hardray} for the construction of weighted map.
For perspective images, we additionally consider the effects of wide-angle field-of-view and assigning a sampling weight inverse to the actual radii of each pixel footprint, which better matches the pixel coverage in 3D world.
The effectiveness of weighted samples can be seen in \cref{fig:hardray2}.

\begin{figure}[t]
    \centering
    \includegraphics[width=\linewidth]{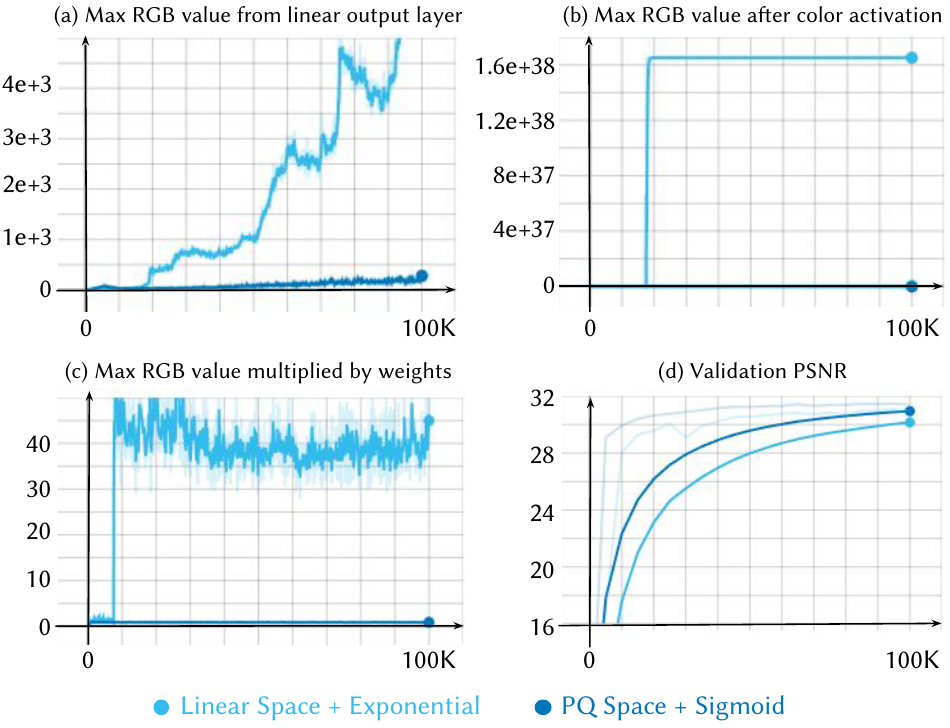}
    \caption{\label{fig:pq-analysis}%
        \textbf{RGB value analysis.}
        These curves show:
        (a) the maximum RGB value obtained from the linear output layer of the color network,
        (b) the maximum RGB value after the color activation function,
        (c) the point-wise color multiplied by the integrated density weight for final color composition, and
        (d) the validation PSNR showing the rendering fidelity.
        The two blue lines represent:
        (1) the baseline using iNGP with linear color space, and exponential activation (light blue), and (2) our PQ color space with sigmoid activation function (dark blue).
    }
\end{figure}

\section{Additional Results and Analysis}

We show a detailed breakdown of our complete quantitative ablation results across all 11 Eyeful Tower datasets in \cref{tab:full_results}.

\subsection{Learning of linear RGB space}

\Cref{fig:pq-space} visualize the converted PQ color space, where both very bright and dark regions are properly preserved in the converted space.
\Cref{fig:pq-analysis} plots the RGB values produced by model linear output layer, activation function, as well as the weighted value used for final volumetric integration during the training process.
It is clear to see that directly learning on linear RGB space with safe exponential function may cause the unstable training where the maximum value predicted from model outputs keeps growing and fluctuated all the way.
In contrast, we need not worry about all these issues in PQ RGB space as we are already work on the bounded PQ space in range [0,1], which gives pretty stabilized learning with commonly used sigmoid activation function.

\subsection{Effects of LOD}

\begin{figure}[t]
	\centering
	\includegraphics[width=\linewidth]{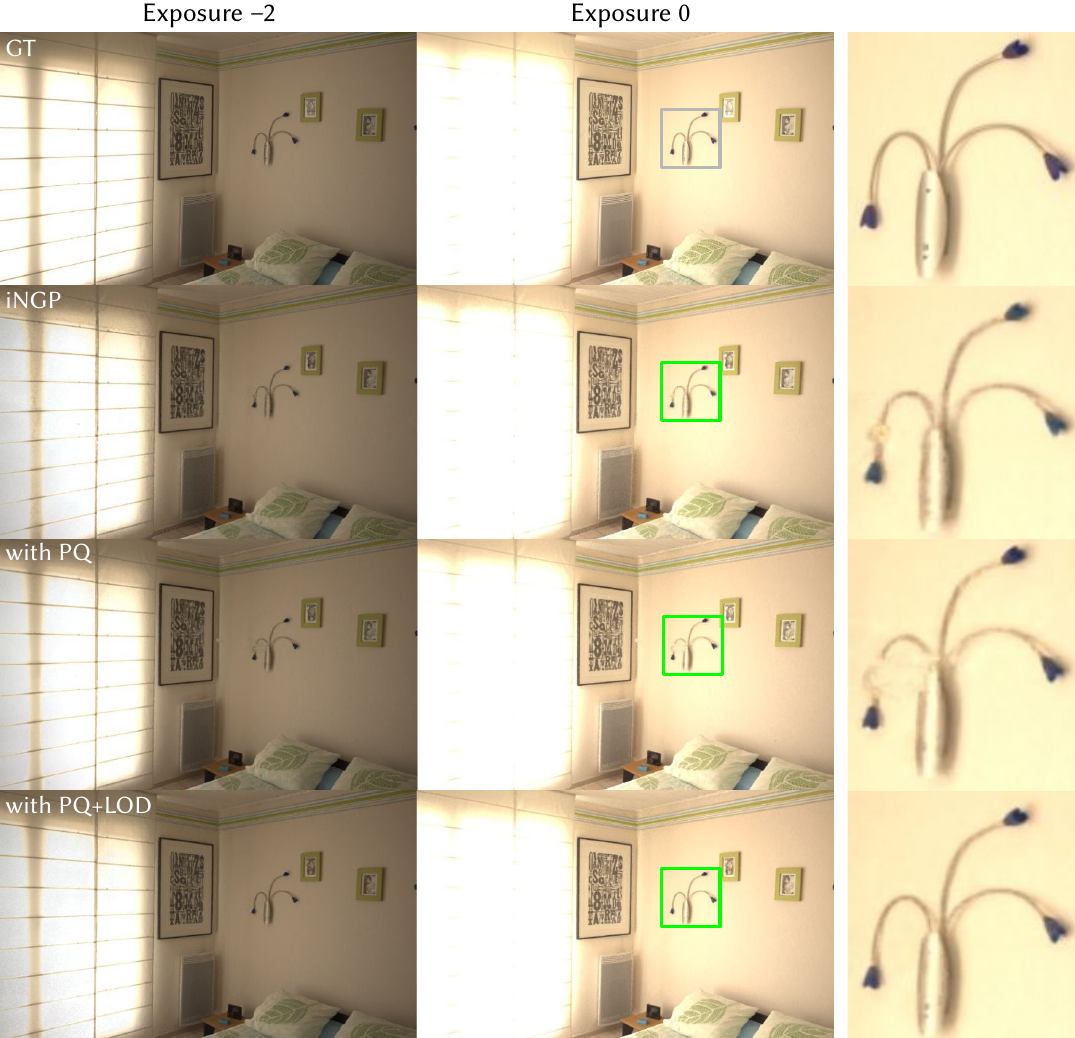}
	\caption{\label{fig:ablation-inria}%
		\textbf{Ablation comparisons on raw Inria dataset \cite{PhiliMGD2021}.}
		The highlighted patches clearly show that iNGP w/ PQ + LOD better preserves the geometry and details compared to the ablated baselines.
		We also adjust the exposure values to adapt to the bright areas around the window.
	}
\end{figure}

\Cref{fig:ablation-inria} shows the two ablated modules on Inria dataset.
One commonly observed advantage of using LOD feature is its improvements on revealing fine details.
We conjecture that by dynamically masking out high-resolution grid features, the model encourages these high-frequency features to only be used for rendering contents with close observations and fine details.

\subsection{Ablation on pruning strategy}

\Cref{fig:ablate-pruning} shows the comparison on different pruning strategies.
As `history pruning' only considers stochastically sampled points visited during the pruning period, it is unlikely to visit all the voxels during the updating period, leading to numerous holes in the obtained occupancy grids.
The quality of `grid pruning' commonly depends on the number of samples placed within each voxel.
The estimation accuracy can get improved with increased number of samples yet at the cost of large computation expenses.
Furthermore, as these samples are usually evenly places for robustness, it can rarely matches with surface points, leading to box-like artifacts in the obtained geometry.
Our joint training combines the merits of each method and achieves accurate pruning results with limited computing budges (4 points for each voxel for grid pruning).

\begin{figure}[b]
	\centering
	\includegraphics[width=\linewidth]{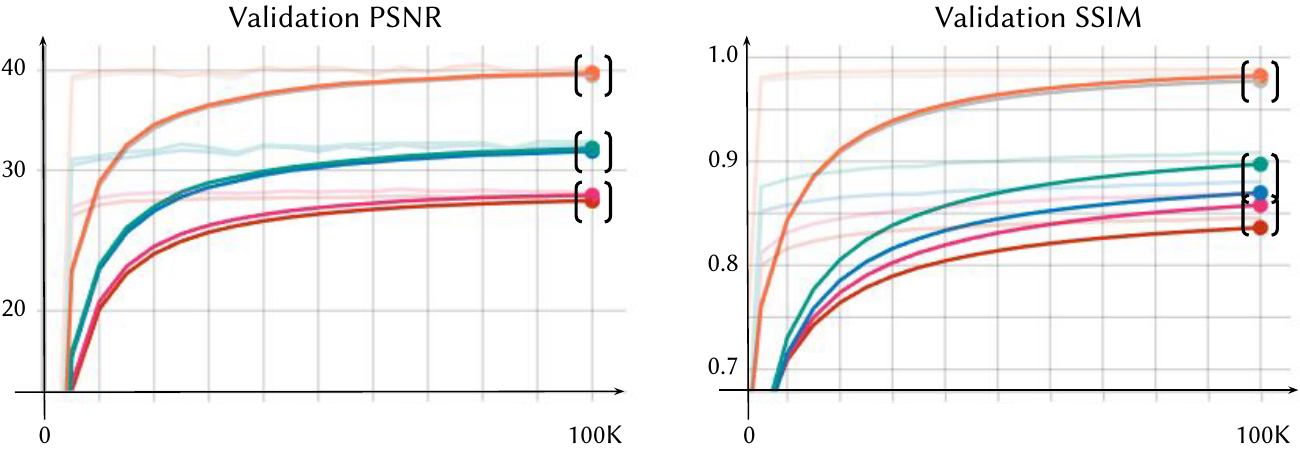}
	\caption{\label{fig:1k-2k}%
	    \textbf{1K \& 2K training.}
	    The learning curves show the comparative validation results between training on 1K \& 2K resolution datasets.
	    The black brackets include the pair of experiments for a same scene with 1K and 2K version.
	    The quantitative metrics are similar without significant drop when adapting to higher resolution images, which shows the potential of using higher-resolution images for training sufficiently long.
	}
\end{figure}

\begin{figure*}[p]
	\centering
	\includegraphics[width=\linewidth]{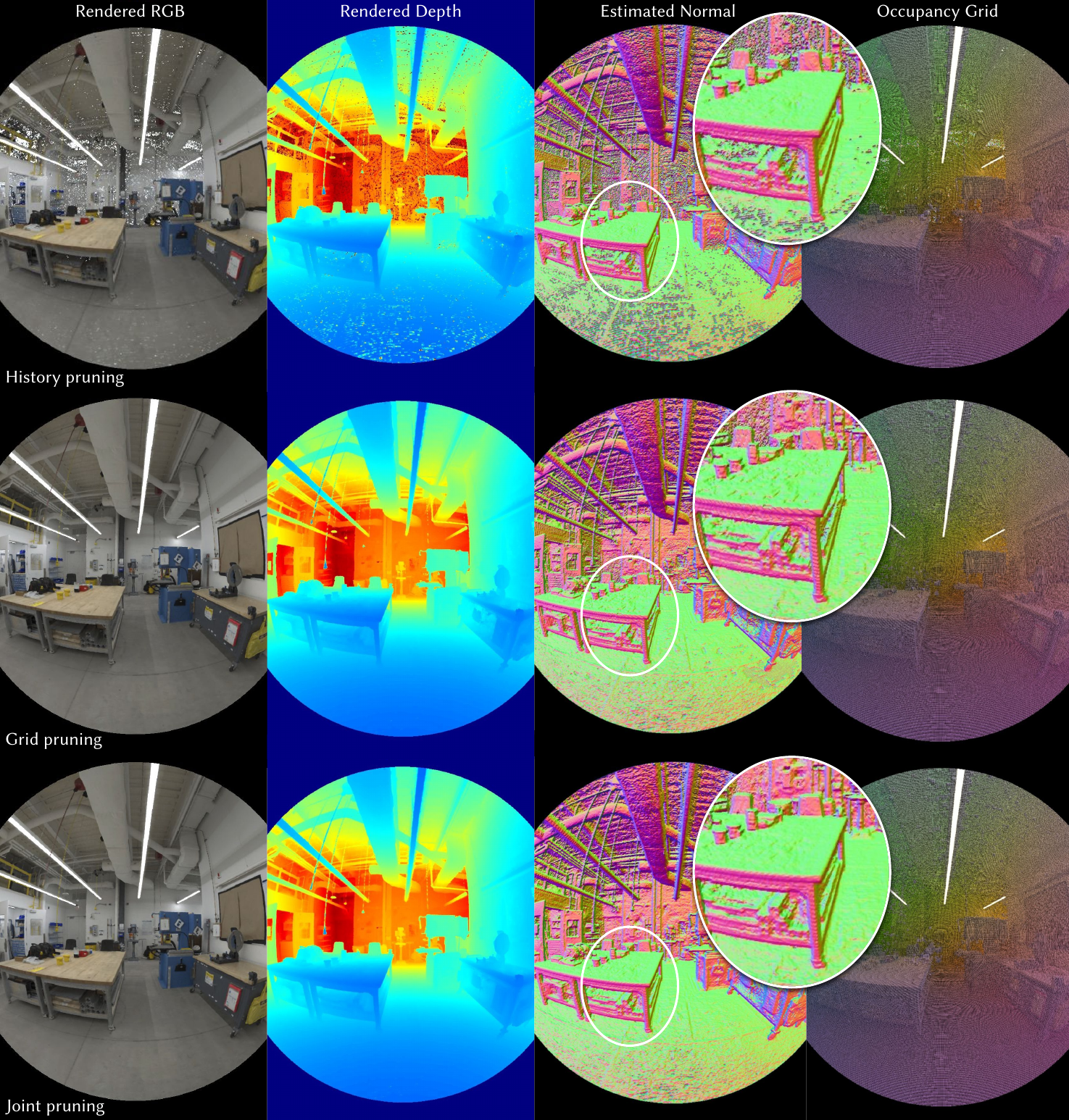}
	\caption{\label{fig:ablate-pruning}%
        \textbf{Ablation on pruning strategy.}
        From top to bottom, we compare three alternative pruning strategies.
        `Joint pruning' (bottom) leverages the advantage of both `history pruning' (top) and `grid pruning' (middle) by placing important sample points observed during training and also densely evaluating voxel grids with sufficient coverage.
        The obtained occupancy grid shown on the right is clean and accurate, and the derived depth map and normal map indicate the well-preserved geometry compared to each individual strategy.
	}
\end{figure*}

\begin{figure*}
	\centering
	\includegraphics[width=\linewidth]{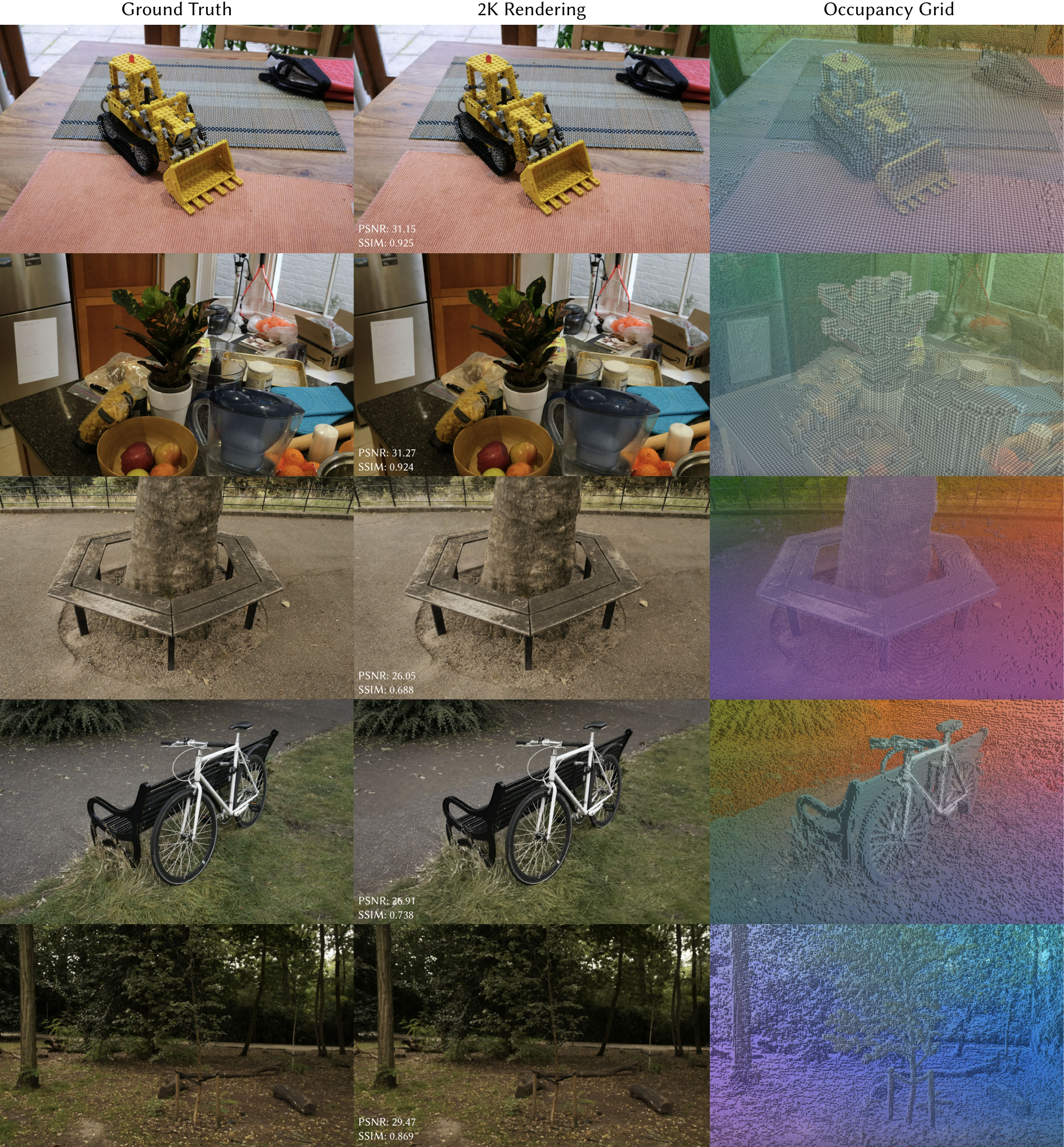}
	\caption{\label{fig:mipnerf360}%
		Additional results on Mip-NeRF 360 scenes \cite{BarroMVSH2022}, trained on 2K resolution images for 50K iterations. (Best zoom in to investigate details.)
	}
\end{figure*}

\begin{figure*}
	\centering
	\includegraphics[width=\linewidth]{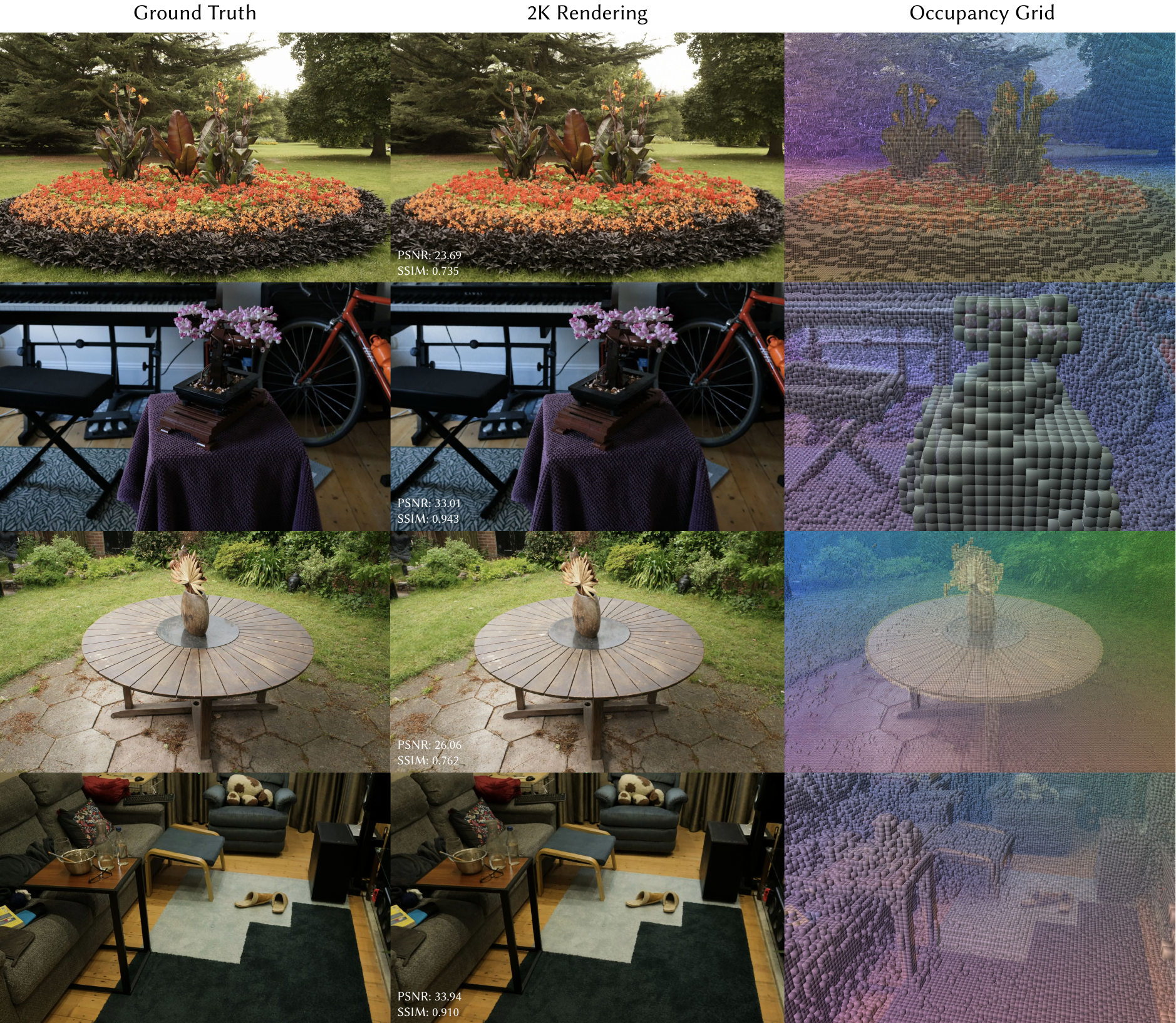}
	\caption{\label{fig:mipnerf360-2}%
		Additional results on Mip-NeRF 360 scenes \cite{BarroMVSH2022}, trained on 2K resolution images for 50K iterations. (Best zoom in to investigate details)
	}
\end{figure*}

\begin{figure*}
	\centering
	\includegraphics[width=\linewidth]{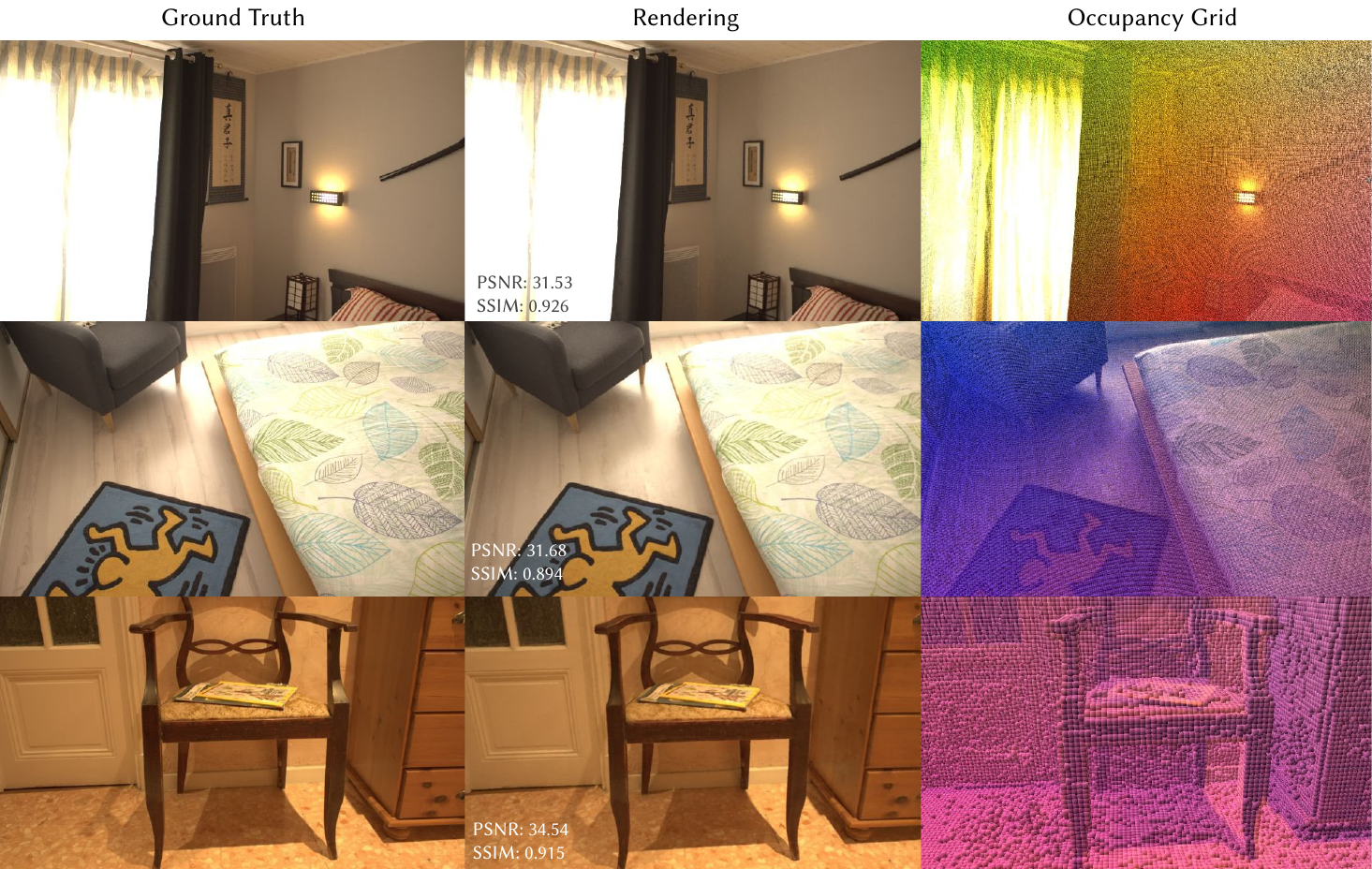}
	\caption{\label{fig:inria}%
		Additional results on Inria scenes \cite{PhiliMGD2021}, trained on 1K resolution images for 100K iterations. (Best zoom in to investigate details)
	}
\end{figure*}

\subsection{Mip-NeRF 360 \& Inria datasets}

We additionally tested on Mip-NeRF 360 dataset \cite{BarroMVSH2022} with our LOD and pruning designs.
\Cref{fig:mipnerf360} and \cref{fig:mipnerf360-2} show all the 9 scenes used in \citet{BarroMVSH2022}.
The models are trained at 2K resolution, with properly recovered fine details and accurate occupancy grids.
\citet{PhiliMGD2021} provide scenes with captured raw images.
\Cref{fig:inria} shows three scenes trained with HDR inputs and PQ color space.

\subsection{High-resolution rendering}

\Cref{fig:1k-2k} shows the learning curve of 1K and 2K training results.
The validation PSNR for each 1K \& 2K pair is generally close, with slight drop in SSIM metric.
\mainorsup{\Cref{fig:4k-render}}{Figure 9 in the main paper} shows an example trained with 4K resolution with fine-grained details.

\subsection{Handling per-image variations}

\begin{figure}[t]
	\centering
	\includegraphics[width=\linewidth]{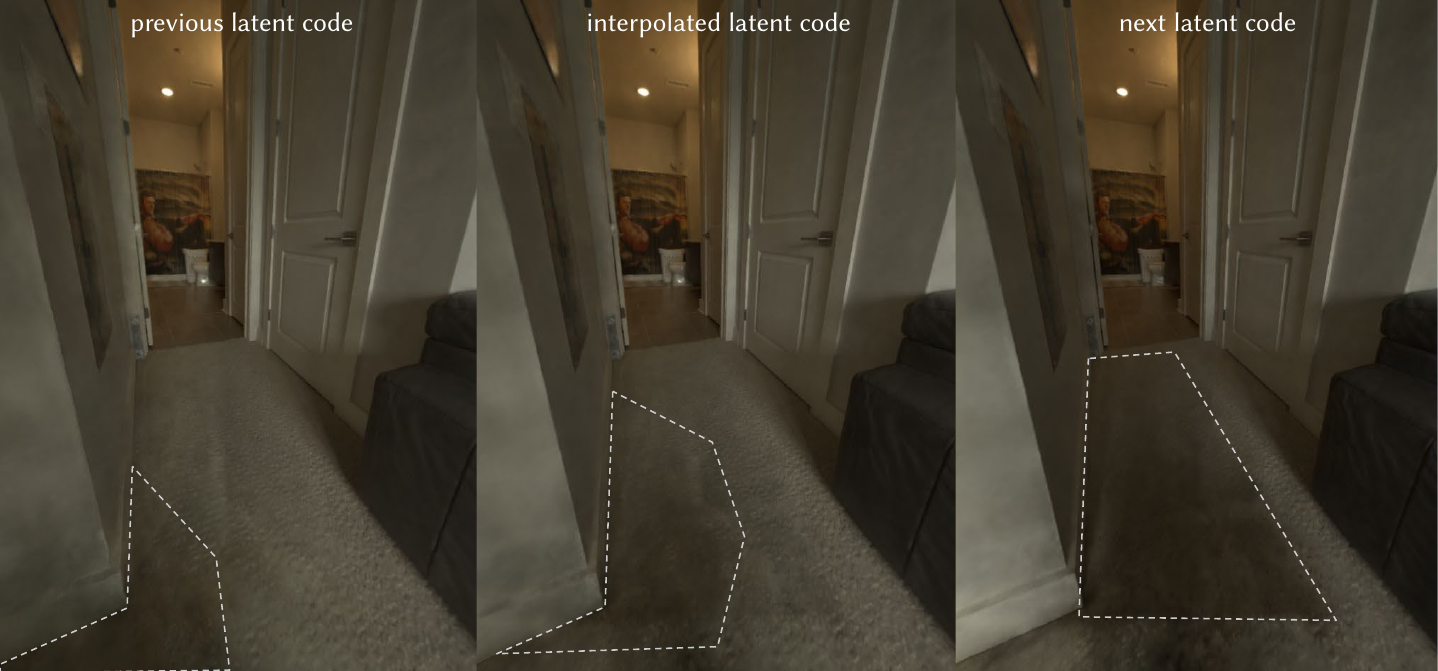}
	\caption{\label{fig:latent}
        \textbf{Latent code condition.} We try to use latent code to explain away shadow issues.
		The image is rendered at the interpolate frame between two adjacent capture timestamps.
		From left to right we show results of using the learned latent code from previous frame, interpolated latent code, and latent code from next frame.
		While showing the tendency of moving the shadows smoothly (highlighted in white polygons), the overall appearance remains noisy with dark floats.
	}
\end{figure}

\begin{figure}[t]
	\centering
	\includegraphics[width=\linewidth]{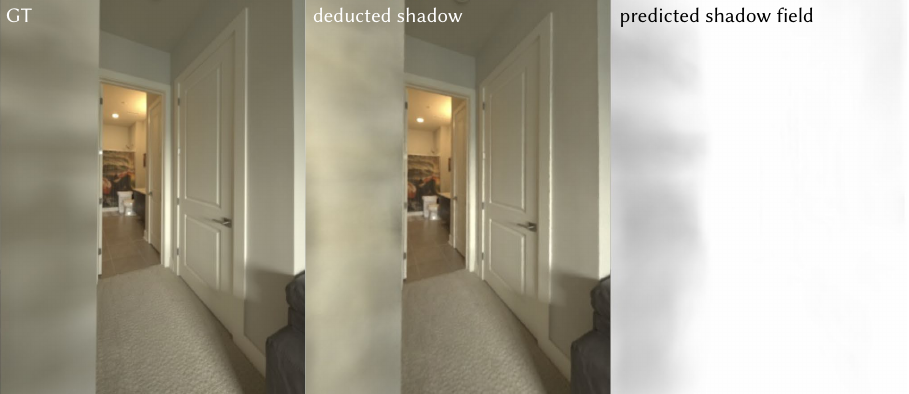}
	\caption{\label{fig:latent2}
        \textbf{Shadow Fields.} We implemented a shadow field \cite{WuZTCO2022} to explain shadows with per-point predicted attenuation.
        We found it helpful to use low-frequency grid feature only for shadow field prediction.
        While the predicted shadow field looks reasonable in general, the accuracy is not sufficient to properly compensate for the affected appearance.
	}
\end{figure}

To explain per-image appearance variations, a latent code is commonly attached to each training image following the practice of \citet{MartiRSBDD2021}.
One specialty of our captured data is that instead of using per-image latent code, we can consider using per-frame latent code (shared by 22 cameras at a same time) as a stronger regularization constraint.
During our capture process, the outdoor lighting conditions can change slightly, and the moving people and capture rig can cast annoying shadows sometimes.
It still remains an open question for us how to deal with these shadows effectively, as we found that using the interpolated latent code (\cref{fig:latent}) or modeling with shadow field \cite{WuZTCO2022} explicitly (\cref{fig:latent2}) can only lead to sub-optimal solutions.
This becomes an extreme challenging scenario when we only have few images for each observation locations while most of them are cast by shadows.

\begin{table}[tb]
\caption{ \label{tab:nerfstudio-table}
	Test results on selected datasets using nerfstudio \cite{TanciWNLYKWKASAMK2023}.
	PSNR/SSIM/LPIPS and FPS are reported here.
}
\sffamily
\begin{tabular}{lllll}
\toprule
                           & PSNR  & SSIM & LPIPS & FPS  \\\midrule
\textsc{seating\_area}     & 30.01 & 0.891 & 0.140  & 0.517 \\
\textsc{workshop}          & 26.85 & 0.849 & 0.277  & 0.570 \\
\textsc{office2}           & 26.60 & 0.939 & 0.127  & 0.595 \\
\textsc{office\_view1}     & 27.01 & 0.804 & 0.382  & 0.374 \\
\textsc{riverview}         & 26.66 & 0.817 & 0.260  & 0.345 \\\bottomrule
\end{tabular}
\end{table}

\subsection{Nerfstudio results (nerfacto)}

Pure MLP-based NeRF methods have difficulty in scaling up due to slow training and limited model capacity.
An alternative baseline we considered is the versatile nerfstudio tool \cite{TanciWNLYKWKASAMK2023}.
We test a subset of our dataset and use nerfstudio for training and evaluations.
We use the integrated nerfacto model, trained on 2K image for 100K iterations.
The parameter setting for hash grid matched with our model ($128 \cdot 1.4^{15}$), and leave other components with default configurations.
The poses are processed with the XML camera pose file produced by Agisoft Metashape.
We use the fisheye lens model for Eyeful Tower v1 images (with cropped black borders), and perspective model for Eyeful Tower v2 images, which are supported directly by nerfstudio.

\begin{figure}
	\centering
	\includegraphics[width=\linewidth]{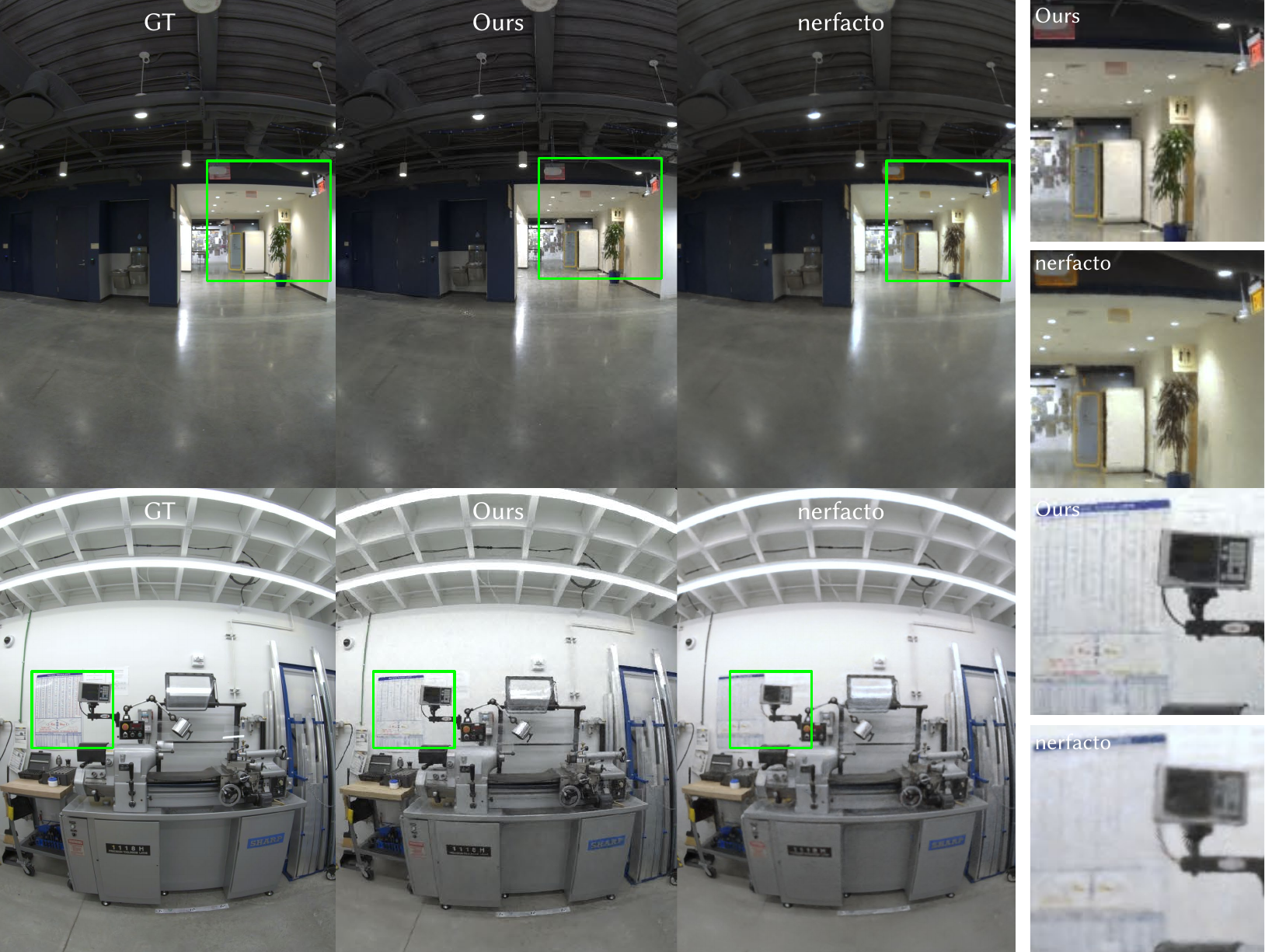}
	\caption{\label{fig:nerfstudio-result}
        \textbf{Qualitative comparison between ours and nerfacto.}
        While the nerfacto model tends to produce scenes with smooth geometry and visuals, our model renders finer details in terms of correct color and high-frequency details, as visible in the zoomed-in patches.
	}
\end{figure}

\cref{tab:nerfstudio-table} and \cref{fig:nerfstudio-result} show the quantitative and qualitative results for nerfstudio nerfactor model trained with sRGB spaces.
We found that they can handle far-field well in unbounded scenes and can capture most details in the scene, yet commonly miss the detailed textures such as those on the carpets and floor.
Note that our results  shown in \cref{fig:nerfstudio-result} are trained on HDR and converted to sRGB space, where we can fairly compare the rendered visual quality.

\section{Additional Renderer Details}

\subsection{Design of our 20-GPU Workstation Machine}
\label{sec:turtle-design}

The design of our custom 20-GPU rendering workstation was driven by the following goals:

\begin{itemize}\setlength\itemsep{.35em}
\item
\textbf{Single CPU.}
Our early experiments revealed higher stability for the Oculus VR runtime with single-socket computers than with dual-socket machines.
Additionally, programming for dual-socket machines requires special considerations, e.g. when crossing NUMA domains.
To maximize stability and minimize programming difficulty, we require a single CPU.

\item
\textbf{16+ direct-attached dual-slot GPUs.}
The GPUs should be available to programs just as the typical 2–4 are on workstations, \ie, without network access or special cluster management software, as a typical render farm would have.
The GPUs should be approximately equivalent to desktop Nvidia RTX 3090 cards.
This, combined with the previous goal, should enable applications written for our multi-GPU workstations to fully utilize the machine with no code changes.

\item
\textbf{Windows 10 OS.}
The Oculus VR stack only works on Windows operating systems.
Using Windows 10 (instead of e.g. Windows 11 or Windows Server) allows the machine to more closely match our development workstations.

\item
\textbf{Mobile and quiet.}
The machine should live inside a movable enclosure that can fit through a standard 32" door, and be quieter than 55\,dB within it.
This enables the machine to be taken to conferences and for demos to be given in the same room.

\end{itemize}

\noindent
A thorough survey of commercial options found no solutions which satisfy all above requirements.
Many vendors offer workstations or servers with 8 GPUs, but nearly all use two CPU sockets.
A few vendors offer 10–16 GPU servers, but these are typically limited to single-slot GPUs, and always use two sockets.
Thus, we build our own solution.
This system is housed in a USystems Edge 3 sound-dampening rack that offers 30\,dB of noise reduction.

\subsection{Efficient Level-of-Detail Rendering}

\mainorsup{\Cref{sec:lod}}{Section 4.2 in the main paper} described the advantages that level-of-detail rendering can have on image quality.
However, our LOD-based masking strategy can also improve rendering performance.
Kernel profiling measurements revealed that substantial time per frame is being spent waiting for features to be sampled from the multi-resolution hash grid, with the largest portion of the time being taken by the finest feature layers.
This is likely due to the hash grid storage that leads to highly incoherent memory accesses.
Coarser levels are stored within a dense linear array, and do not suffer as much from this issue, though the memory layout is still suboptimal for spatial coherence.
Our LOD-based masking strategy removes the need to sample many of these expensive hash-grid features, and thus decrease rendering time, as we're able to conditionally replace the finest sampled levels with zeros.

\end{document}